\documentclass[journal]{IEEEtran}
\ifCLASSINFOpdf
\else
\fi
\usepackage{graphicx}
\usepackage{amsmath}
\usepackage{amssymb}
\usepackage{algorithm}
\usepackage{algpseudocode}
\usepackage{multirow}
\usepackage{booktabs}

\begin{document}
%
\title{AttriBE: Quantifying Attribute Expressivity in Body Embeddings for Recognition and Identification}
%
%
%

\author{Basudha Pal,~\IEEEmembership{Graduate Student Member,~IEEE,}
        Siyuan Huang,~\IEEEmembership{Graduate Student Member,~IEEE,}
        Anirudh Nanduri,~\IEEEmembership{Graduate Student Member,~IEEE,}
        Zhaoyang Wang,~\IEEEmembership{Graduate Student Member,~IEEE,}
        and~Rama Chellappa,~\IEEEmembership{Life~Fellow,~IEEE}

\thanks{Basudha Pal, Siyuan Huang, and Rama Chellappa are with the Department of Electrical and Computer Engineering,
Johns Hopkins University, Baltimore, MD 21218 USA.}
\thanks{Rama Chellappa is also with the Department of Biomedical Engineering,
Johns Hopkins University, Baltimore, MD 21218 USA.}
\thanks{Anirudh Nanduri is with the Department of Electrical and Computer Engineering,
University of Maryland, College Park, MD 20742 USA.}
\thanks{Zhaoyang Wang with the Department of Computer Science,
Johns Hopkins University, Baltimore, MD 21218 USA.}
\thanks{e-mail: bpal5@jhu.edu; shuan124@jhu.edu; rchella4@jhu.edu}
}

%
%

\markboth{Journal of \LaTeX\ Class Files,~Vol.~14, No.~8, August~2015}%
{Shell \MakeLowercase{\textit{et al.}}: Bare Demo of IEEEtran.cls for IEEE Journals}
%



\maketitle

\begin{abstract}
Person re-identification (ReID) systems that match individuals across images or video frames are essential in many real-world applications. However, existing methods are often influenced by attributes such as gender, pose, and body mass index (BMI), which vary in unconstrained settings and raise concerns related to fairness and generalization. To address this, we extend the notion of \textit{expressivity}, defined as the mutual information between learned features and specific attributes, using a secondary neural network to quantify how strongly attributes are encoded. Applying this framework to three transformer-based ReID models on a large-scale visible-spectrum dataset, we find that BMI consistently shows the highest expressivity in deeper layers. Attributes in the final representation are ranked as $\text{BMI} > \text{Pitch} > \text{Gender} > \text{Yaw}$, and expressivity evolves across layers and training epochs, with pose peaking in intermediate layers and BMI strengthening with depth. We further extend the analysis to cross-spectral person identification across infrared modalities including short-wave, medium-wave, and long-wave infrared. In this setting, pitch becomes comparable to BMI and attribute trends increase monotonically across depth, suggesting increased reliance on structural cues when bridging modality gaps. Overall, the results show that transformer-based ReID embeddings encode a hierarchy of implicit attributes, with morphometric information persistently embedded and pose contributing more strongly under cross-spectral conditions.
\end{abstract}

\begin{IEEEkeywords}
Mutual Information Neural Estimation, Person Re-identification, Explainability.
\end{IEEEkeywords}

\section{Introduction}

\let\thefootnote\relax\footnotetext{This article extends our 2025 IEEE International Joint Conference on Biometrics (IJCB) paper with new cross-spectral analysis and expanded theoretical validation.}

Deep neural networks trained for biometric recognition are optimized to encode identity-discriminative information, yet they frequently capture \emph{unintended} latent attributes that are not explicitly supervised. In face recognition, identity embeddings have been shown to cluster according to demographic variables such as gender while also encoding nuisance factors including pose, age, and illumination \cite{hill2019deep, nagpal2019deep, parde2017face}. Such latent encoding can influence recognition accuracy and operational behavior under varying capture conditions \cite{givens2013introduction, lee2014generalizing}. Systematically characterizing which auxiliary attributes are embedded in learned representations, and the strength with which they are encoded, is therefore essential for interpretability, robustness analysis, and bias diagnosis in biometric systems.

To address this question, Dhar \emph{et al.} introduced the notion of \emph{expressivity} for face recognition, quantifying the relationship between deep network features and semantic attributes \cite{dhar2020attributes}. Building on this concept, we extend expressivity analysis to \emph{person re-identification (ReID)} with the goal of understanding how body-related attributes are represented within models trained primarily for identity matching. Person ReID plays a central role in smart-city infrastructure, surveillance analytics, and public safety \cite{behera2020person, khan2024deep}, as well as in autonomous perception systems for pedestrian detection and tracking \cite{camara2020pedestrian, wong2020identifying}. The task requires reliable identity matching across non-overlapping camera views under substantial variations in pose, appearance, and environment \cite{zheng2015scalable}. Significant advances have been achieved through both image-based and video-based modeling, including temporal aggregation and robustness to clothing or appearance variation \cite{gu2019temporal, gu2022clothes}.

Despite these advances, modern ReID systems can unintentionally encode auxiliary body attributes beyond identity, motivating principled post-hoc analysis of internal feature representations. Recent work has demonstrated the presence of attribute information in body-recognition embeddings using downstream probing techniques such as logistic regression \cite{metz2025dissecting}. However, probing accuracy reflects only \emph{predictability} and provides an indirect proxy for representational dependence, potentially missing nonlinear or non-Gaussian relationships. In contrast, information-theoretic analysis enables direct quantification of statistical dependency between learned features and attributes. Accordingly, we adopt Mutual Information Neural Estimation (MINE) \cite{belghazi2018mutual}, grounded in classical information theory \cite{cover1999elements}, to measure how much attribute-related information is intrinsically encoded within deep biometric representations.

Our prior study accepted at the International Joint Conference on Biometrics (IJCB) introduced this information-theoretic expressivity framework for transformer-based person ReID and analyzed three state-of-the-art architectures namely, SemReID, DCFormer, and PFD, using visible-spectrum imagery from the Broad Range Identification at Altitude and Range (BRIAR) dataset \cite{pal2025quantitative, cornett2023expanding}. That work provided, to our knowledge, the first systematic quantification of body-attribute encoding in deep person-recognition embeddings through mutual information (MI) analysis. However, the investigation was restricted to a single sensing modality and did not examine how attribute encoding behaves under spectral domain shift.

In operational biometric environments, recognition systems must function across heterogeneous sensing conditions, particularly visible (VIS) and infrared (IR) spectra, where sensing physics, appearance statistics, and noise characteristics differ substantially. Cross-spectral person identification, often referred to as Visible–Infrared Person Re-Identification (VI-ReID) has historically been studied using single infrared bands such as near-infrared (NIR) or long-wave infrared (LWIR), frequently relying on image translation or modality synthesis between VIS and IR domains. However, such approaches do not capture the full complexity of multi-domain infrared sensing.

The IARPA Janus Benchmark Multi-Domain Face (IJB-MDF) dataset introduces a cross-spectral recognition protocol requiring identity matching between VIS imagery and multiple infrared domains, including short-wave infrared (SWIR), medium-wave infrared (MWIR), and long-wave infrared (LWIR). Recent work on multi-domain biometric recognition using body embeddings demonstrated that body-based representations can outperform face-based embeddings for cross-spectral identification in MWIR and LWIR conditions, and established the first benchmark results using the SemReID architecture within the MDF protocol. This study further provided insights into inter-domain infrared relationships, transferability of VIS-pretrained models, the importance of local semantic body features, and effective training strategies for limited multi-domain data \cite{nanduri2025multi}.

Motivated by these findings, the present work introduces the first mutual-information–based expressivity analysis of body embeddings across spectral domains under the MDF recognition protocol. The study centers on the SemReID foundation model \cite{huang2023self}. This choice is supported by two complementary observations: prior comparative evaluation established SemReID as providing strong rank-accuracy performance and stable representation quality among contemporary transformer-based ReID architectures \cite{pal2025quantitative,huang2023self}, while subsequent work on multi-domain biometric recognition using body embeddings demonstrated its effectiveness for cross-spectral identity matching within the MDF framework \cite{nanduri2025multi} and on the LLCM benchmark \cite{zhang2023diverse}. Building on this foundation, we investigate whether the attribute-encoding hierarchies observed in visible-spectrum person re-identification remain consistent, shift, or intensify across SWIR, MWIR, and LWIR domains, providing new insight into the robustness, fairness, and deployability of biometric recognition under real-world sensing conditions. The principal contributions of this work are summarized as follows:

\begin{itemize}
    \item We introduce the first MI-based framework for analyzing body-attribute encoding in transformer-based person re-identification representations, enabling layer-wise characterization of semantic attribute dependence within a large-scale self-supervised Vision Transformer (ViT) foundation model, SemReID \cite{huang2023self}.
    
    \item Using Mutual Information Neural Estimation (MINE) \cite{belghazi2018mutual}, we quantify and rank attribute encoding strength in deep feature spaces, revealing stable hierarchical structure in how body-related attributes are represented across network depth and spectral sensing conditions.
    
    \item We extend visible-spectrum analysis on the BRIAR benchmark \cite{cornett2023expanding} to the first cross-spectral expressivity study under the MDF recognition protocol \cite{kalka2019iarpa, nanduri2024template, nanduri2025multi}, providing new insight into how feature–attribute dependencies evolve across visible, SWIR, MWIR, and LWIR domains relevant to real-world biometric deployment.
\end{itemize}

\section{Related Works}
\subsection{Bias and interpretability in biometric recognition}
Bias and interpretability in biometric systems have been extensively studied, particularly in face recognition and related domains. Prior work has shown that identity representations often encode demographic and nuisance attributes, influencing both embedding geometry and recognition behavior \cite{hill2019deep, nagpal2019deep, parde2017face}. More broadly, bias in recognition pipelines has been systematically analyzed, with several works proposing mitigation strategies for demographic and attribute-based biases \cite{schwemmer2020diagnosing, dhar2021pass, siddiqui2022examination, pal2024gamma, pal2024diversinet}. These findings highlight the need for principled methods to quantify attribute leakage and disentangle identity-relevant information from confounding factors.

\subsection{Post-hoc interpretability methods}
A large body of work studies trained models using post-hoc interpretability techniques. Concept-based approaches such as TCAV measure sensitivity to user-defined concepts by learning Concept Activation Vectors via linear classifiers \cite{kim2018interpretability}. While effective for discrete attributes (e.g., color or texture), TCAV is less suitable for continuous or pervasive attributes such as pose or BMI due to the difficulty of defining meaningful negative examples and its reliance on samples from seen classes, limiting applicability in open-set settings. 

Other approaches include layer-wise probing using linear classifiers \cite{alain2016understanding}, influence-based techniques that quantify sensitivity to training data perturbations \cite{koh2017understanding}, and saliency methods that produce spatial explanations via gradient-based localization \cite{selvaraju2017grad, chattopadhay2018grad}. While these methods provide insights into model behavior, they typically do not quantify the amount of attribute information encoded in learned representations. 

Beyond post-hoc analysis, several works have explored interpretability through architectural or generative modifications. Schumann \emph{et al.} \cite{schumann2017person} enriched CNN representations using auxiliary networks, while Myers \emph{et al.} \cite{myers2023recognizing} incorporated both linguistic and non-linguistic cues for identity prediction. Yin \emph{et al.} \cite{yin2019towards} introduced spatial activation diversity losses to improve interpretability, and Kim \emph{et al.} \cite{kim2014bayesian} proposed prototype-based generative models for case-based reasoning. However, as discussed in \cite{kim2018interpretability}, many such approaches are restricted to models trained from scratch and do not readily generalize to deployed networks.

In the context of person re-identification, Chen \emph{et al.} proposed explainability via attribute-guided metric distillation \cite{chen2021explainable}, though this approach relies on additional supervision and is tailored to CNN-based architectures. Overall, while saliency and concept-based methods highlight spatial or semantic relevance, they are limited in capturing diffuse or abstract attributes and do not provide a unified quantitative measure of feature--attribute dependency.

\subsection{Expressivity and information-theoretic dependency measures}
Expressivity provides a principled framework for quantifying how attributes are encoded in learned representations. Dhar \emph{et al.} introduced expressivity in deep face recognition models, revealing a hierarchical structure where attributes such as age, sex, and pose are differentially encoded \cite{dhar2020attributes}. Motivated by information-theoretic perspectives on representation learning \cite{cover1999elements, tishby2015deep}, we quantify feature--attribute dependency using Mutual Information Neural Estimation (MINE) \cite{belghazi2018mutual}, which is well suited for capturing nonlinear relationships between high-dimensional embeddings and both discrete and continuous attributes.

Unlike prior interpretability methods, this formulation enables direct, quantitative comparison of identity-related and auxiliary attribute information across network depth and training stages, providing a unified and architecture-agnostic analysis of representation behavior.

\subsection{Person ReID and cross-domain biometrics}
Person re-identification (ReID) aims to match individuals across non-overlapping camera views under challenging conditions such as variations in illumination, clothing, pose, and occlusion~\cite{zheng2015scalable, gu2022clothes, gu2019temporal, huang2019celebrities}. Extensive efforts have addressed these challenges across diverse settings, including Clothes-Changing ReID (CC-ReID)~\cite{gu2022clothes}, video-based ReID with temporal modeling~\cite{cao2023event, hou2020temporal, yan2020learning, zhang2020multi, wu2022cavit}, unconstrained and long-range recognition scenarios~\cite{cornett2023expanding, liu2024farsight, nikhal2023weakly, nikhal2024hashreid, zhu2024sharc}, and short-term or limited-appearance variation settings~\cite{chen2023beyond, zhang2020multi, wang2018learning, zhu2022pass}. These works collectively emphasize robustness to real-world variability through improved feature learning, temporal aggregation, and domain generalization. Among recent approaches, SemReID~\cite{huang2023self} demonstrates state-of-the-art performance across multiple such scenarios by leveraging self-supervised semantic representations.

Despite this progress, interpretability in ReID systems remains relatively underexplored compared to broader recognition domains such as face recognition, where attribute-level understanding has received significant attention. Recent advances in large-scale and unconstrained benchmarks, such as BRIAR, further highlight the complexity of whole-body recognition under real-world conditions and motivate the need for robust and interpretable representations~\cite{cornett2023expanding, liu2024farsight}. In this work, we extend the evaluation to cross-spectral settings using the IJB-MDF benchmark, which spans visible and multiple infrared domains~\cite{kalka2019iarpa, nanduri2024template}. This multi-domain setting provides a critical testbed to assess whether attribute encoding and leakage behavior remain consistent under significant modality shifts, offering deeper insights into the generalization and robustness of learned representations.

%
\IEEEpeerreviewmaketitle

\begin{figure*}[!htbp]
  \centering
  \includegraphics[width=0.7\linewidth]{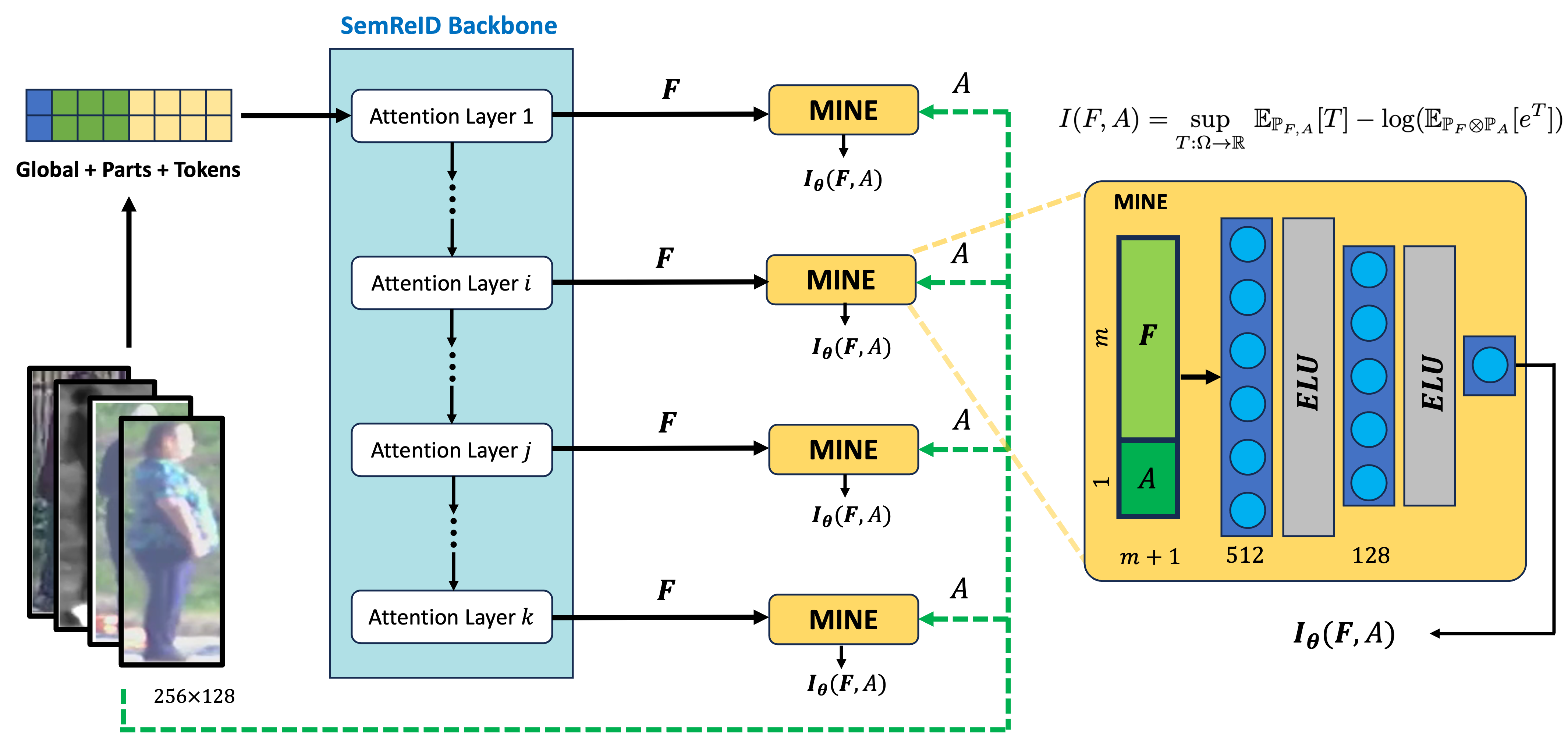}
  \caption{Integration of the MINE block with the ViT-based SemReID \cite{huang2023self} backbone for estimating attribute expressivity in learned body representations. The MINE module employs a two-layer MLP to compute expressivity of \(m\)-dimensional features \(F\). The feature vector is concatenated with the attribute vector \(A\), resulting in an augmented representation of dimension \(m+1\).}
  \vspace{-3mm}
  \label{mainfig}
\end{figure*}

\section{Proposed Method}

The proposed framework, illustrated in Figure~\ref{mainfig}, quantifies the degree to which body recognition models encode attribute-related information in their learned representations. While conventional analyses focus primarily on visible imagery, we extend this formulation to cross-spectral settings by incorporating short-wave (SWIR), medium-wave (MWIR) and long-wave infrared (LWIR) imagery from the IJB-MDF dataset. This unified formulation enables consistent analysis of attribute expressivity across sensing modalities and provides insight into how attribute encoding varies under spectral shifts.

To measure attribute encoding, we employ Mutual Information (MI), which captures statistical dependence between learned features and annotated attributes. Intuitively, MI measures how much knowledge of one variable reduces uncertainty about another. In our context, higher MI indicates that body embeddings contain stronger attribute-related information. This makes MI a natural choice for quantifying attribute expressivity.

Formally, MI between two random variables \(X\) and \(Z\) is defined as

\begin{equation}
\label{eqn1}
I(X ; Z) = \int_{\mathcal{X} \times \mathcal{Z}} 
\log \frac{d \mathbb{P}_{XZ}}{d \mathbb{P}_X \otimes \mathbb{P}_Z} 
d \mathbb{P}_{XZ}.
\end{equation}

This formulation can equivalently be expressed as the Kullback-Leibler divergence between the joint distribution and the product of marginal distributions,

\begin{equation}
\label{eqn2}
I(X ; Z) = D_{KL}\left(\mathbb{P}_{XZ} \| 
\mathbb{P}_X \otimes \mathbb{P}_Z\right).
\end{equation}

This interpretation highlights that MI measures deviation from statistical independence. When features and attributes are independent, MI is zero. Larger values of MI indicate stronger attribute encoding within the learned representations.

\subsection{Problem Setup}

\begin{figure}[t]
  \centering
  \includegraphics[width=0.8\linewidth]{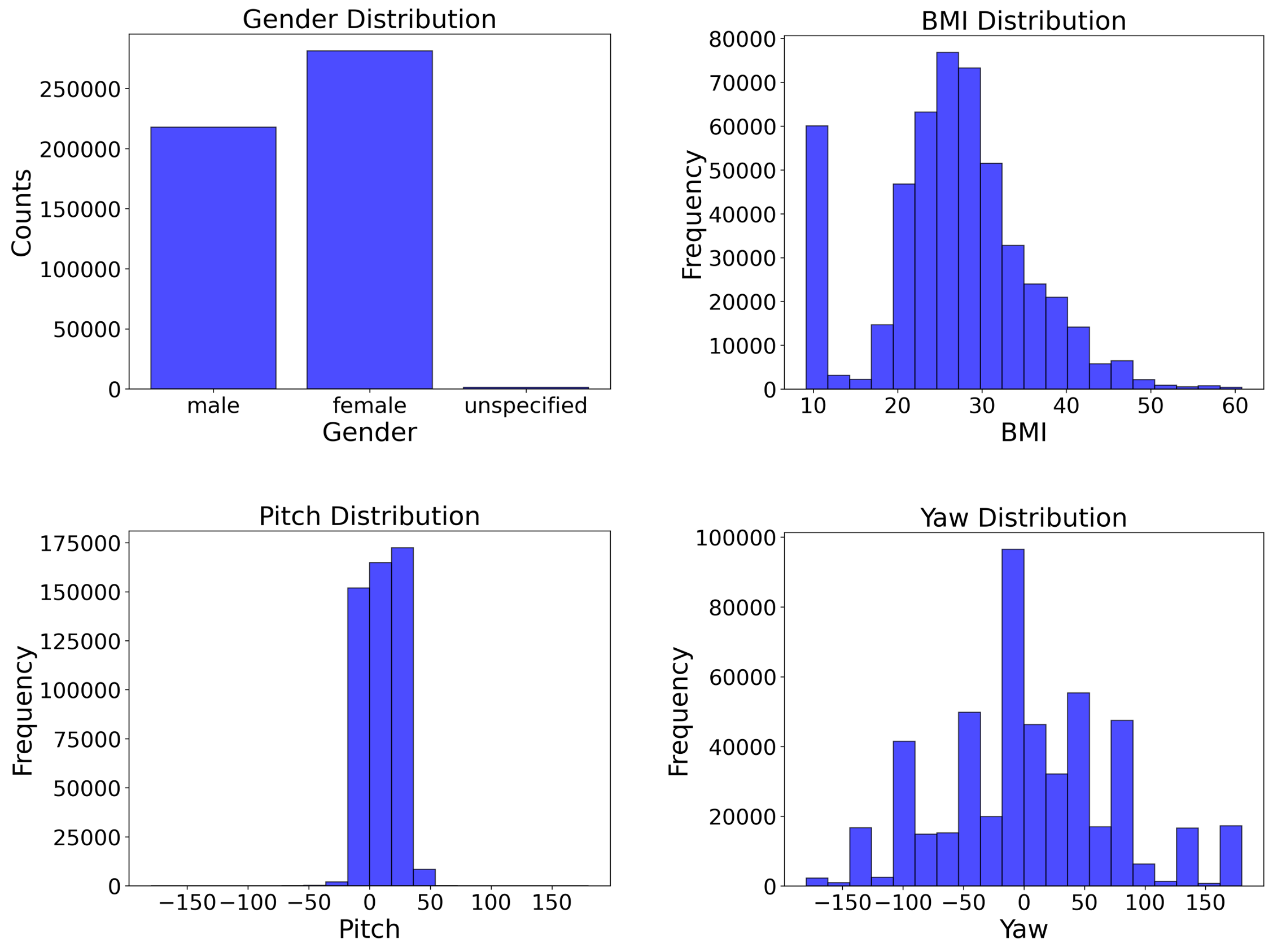}
  \caption{Attribute distribution in the BRIAR dataset, demonstrating sufficient variation for expressivity estimation.}
  \label{attr}
\end{figure}

Our datasets consist of body images captured under varying acquisition conditions and sensing modalities. In addition to visible imagery from BRIAR, we extend the analysis to SWIR, MWIR, and LWIR domains from IJB-MDF. Each image is annotated with identity labels and attributes including gender (\(g\)), height (\(h\)), weight (\(w\)), BMI, pitch (\(p\)), and yaw (\(y\)). These attributes provide complementary physical and pose-related information, enabling comprehensive analysis of attribute encoding across spectral domains.

Let \( \mathbf{F} \in \mathbb{R}^{n \times m} \) denote feature descriptors extracted from a ReID backbone, and let \( \mathbf{A} \in \mathbb{R}^{n \times 1} \) denote the corresponding attribute vector. Each image \(x_i\), regardless of modality (VIS, MWIR, or LWIR), is processed to extract feature descriptors \(f_i\). These descriptors are stacked to form the feature matrix

\[
\mathbf{F} = [f_1, f_2, \dots, f_n]^T.
\]

The attribute vector is concatenated with \(\mathbf{F}\) to obtain an augmented representation

\[
\mathbf{X} = [\mathbf{F} \mid \mathbf{A}],
\]

which jointly encodes both representation and attribute information. This augmented matrix serves as input to the MI estimator.

\subsection{Expressivity of Attributes in Learned Features}

We compute expressivity for four attributes: gender, BMI, pitch, and yaw. Figure~\ref{attr} confirms sufficient variation for reliable estimation. Gender is binary, whereas BMI and pose attributes are continuous. These attributes influence body representations and may be encoded differently across sensing modalities. In particular, infrared imagery lacks color and texture cues present in visible imagery, which may affect how pose and body shape information are represented. Measuring MI across VIS, MWIR, and LWIR therefore provides insight into cross-spectral attribute encoding.

To estimate MI in high-dimensional feature spaces, we adopt the Mutual Information Neural Estimator (MINE) \cite{belghazi2018mutual}. Following the information bottleneck perspective of \cite{tishby2015deep}, this allows us to quantify how information about attributes is retained within learned representations. MINE approximates MI using a neural network \(T_\theta\), yielding

\begin{equation}
I_{\theta}(\mathbf{F}, \mathbf{A}) =
\sup_{\theta \in \Theta}
\mathbb{E}_{P_{FA}}[T_\theta(f,a)]
-
\log
\mathbb{E}_{P_F \otimes P_A}
[e^{T_\theta(f)}].
\end{equation}

The first term measures dependence under the joint distribution, while the second term removes bias introduced by marginal statistics. In practice, these expectations are approximated using mini-batches. The joint expectation is computed as

\begin{equation}
\mathbb{E}_{P_{FA}}[T_\theta(f,a)]
\approx
\frac{1}{b}
\sum_{i=1}^{b}
T_\theta(f_i,a_i),
\end{equation}

and the marginal expectation is approximated as

\begin{equation}
\mathbb{E}_{P_F \otimes P_A}
[e^{T_\theta(f)}]
\approx
\frac{1}{b}
\sum_{i=1}^{b}
e^{T_\theta(f_i)}.
\end{equation}

Combining these terms yields the MI lower bound

\begin{equation}
V(\theta) =
\frac{1}{b}
\sum_{i=1}^{b}
T_\theta(f_i,a_i)
-
\log
\left(
\frac{1}{b}
\sum_{i=1}^{b}
e^{T_\theta(f_i)}
\right),
\end{equation}

which is maximized during training. The corresponding loss is defined as

\begin{equation}
L(\theta) = -V(\theta).
\end{equation}

The gradient of this loss is given by

\begin{equation}
\nabla_\theta L(\theta)
=
-
\left(
\mathbb{E}_{P_{FA}}
[\nabla_\theta T_\theta]
-
\mathbb{E}_{P_F \otimes P_A}
[\nabla_\theta e^{T_\theta}]
\right).
\end{equation}

Algorithm~\ref{alg:expressivity} summarizes the overall computation procedure. For each layer and spectral modality, features are extracted and concatenated with attribute vectors. The MINE estimator is initialized multiple times to reduce variance, and the resulting MI estimates are averaged to obtain the final expressivity score. This procedure ensures stable estimation across visible and infrared domains.

\begin{algorithm}[H]
\caption{Computation of attribute expressivity from learned representations}
\label{alg:expressivity}
\begin{algorithmic}[1]
\Require Layer \(L\), images \(I\) (VIS/SWIR/MWIR/LWIR), attribute vector \(\mathbf{A}\)
\Ensure Expressivity
\State Initialize \(E \gets [ ]\)
\State Extract features \(\mathbf{F}\)
\State \(\mathbf{X} \gets [\mathbf{F} | \mathbf{A}]\)
\For{iteration = 1 to \(M\)}
\State Initialize MINE \(T_\theta\)
\State \(e \gets \text{MINE}(\mathbf{X})\)
\State \(E \gets E \cup \{e\}\)
\EndFor
\State Return Average(\(E\))
\end{algorithmic}
\end{algorithm}

\section{Experiments}

\subsection{Datasets and Settings}

\begin{figure}[htbp]
  \centering
  \includegraphics[width=0.7\linewidth]{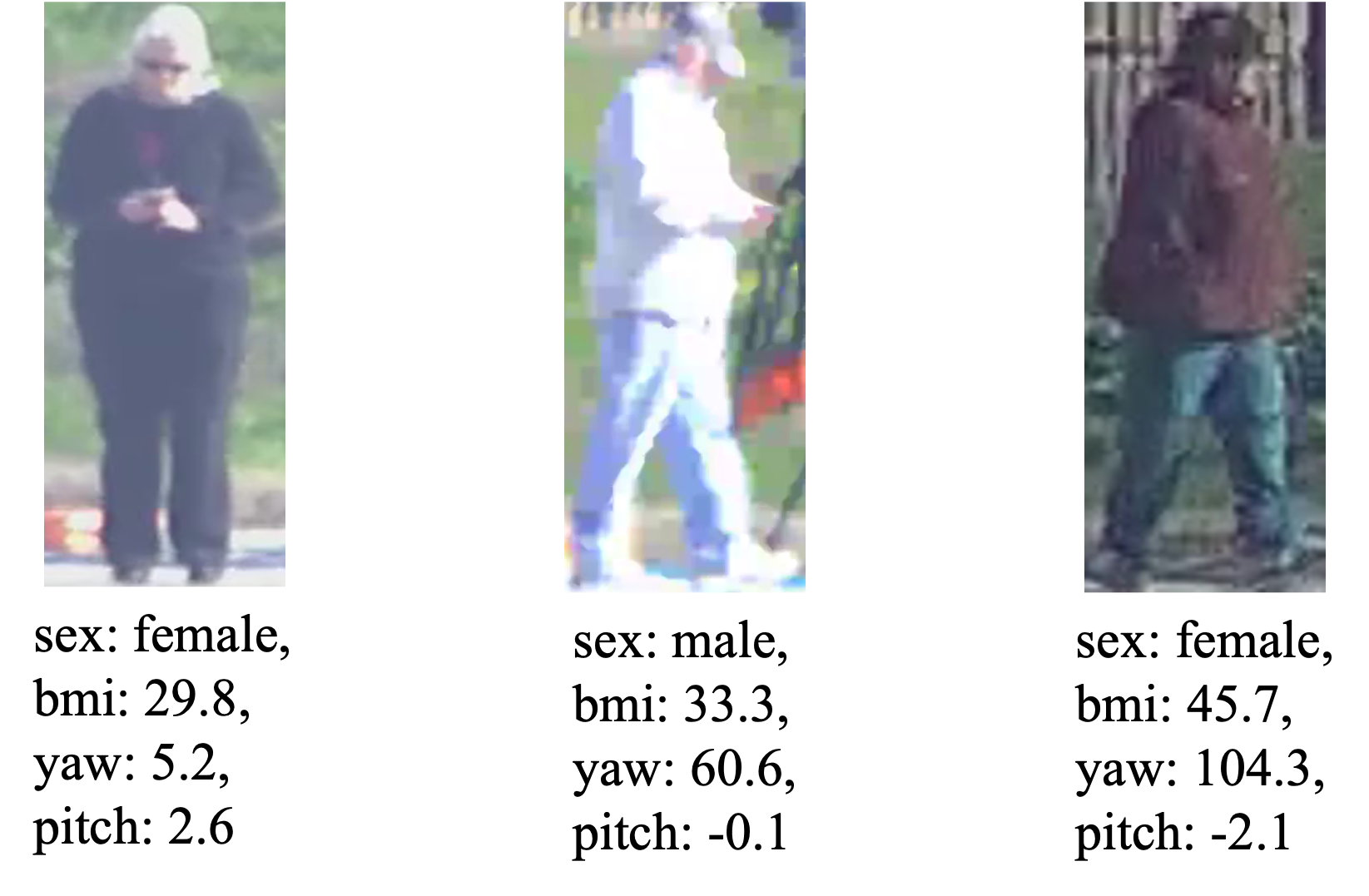}
  \caption{Attribute annotated exemplar images from the BRIAR dataset. All subjects involved provided informed consent for their participation, including the use of their images in research publications and figures.}
  \vspace{-2mm}
  \label{examplesubj}
\end{figure}

We first evaluate attribute expressivity on the BRIAR 1–5 dataset~\cite{cornett2023expanding}, a large-scale unconstrained person re-identification benchmark containing over 1 million images and 40,000 videos captured under diverse real-world conditions. These include varying clothing, long-range acquisition (100m–1km), altitude variations (e.g., UAV), and challenges such as occlusion, motion blur, and atmospheric turbulence. The dataset is organized into five subsets (BRIAR-1 to BRIAR-5) of increasing difficulty with progressively larger identity pools and distractor sets.

From BRIAR, we extract 704,999 frames from 382,229 images and 170,522 videos, corresponding to 2,077 unique identities (887 male and 1,190 female). Each subject is annotated with attributes including gender, height, weight, BMI, and pose (pitch and yaw). Figure~\ref{examplesubj} shows representative examples from our curated subset. The wide variation in acquisition conditions makes BRIAR well suited for analyzing attribute encoding in unconstrained visible-spectrum imagery.

\begin{figure}[htbp]
  \centering
  \includegraphics[width=0.7\linewidth]{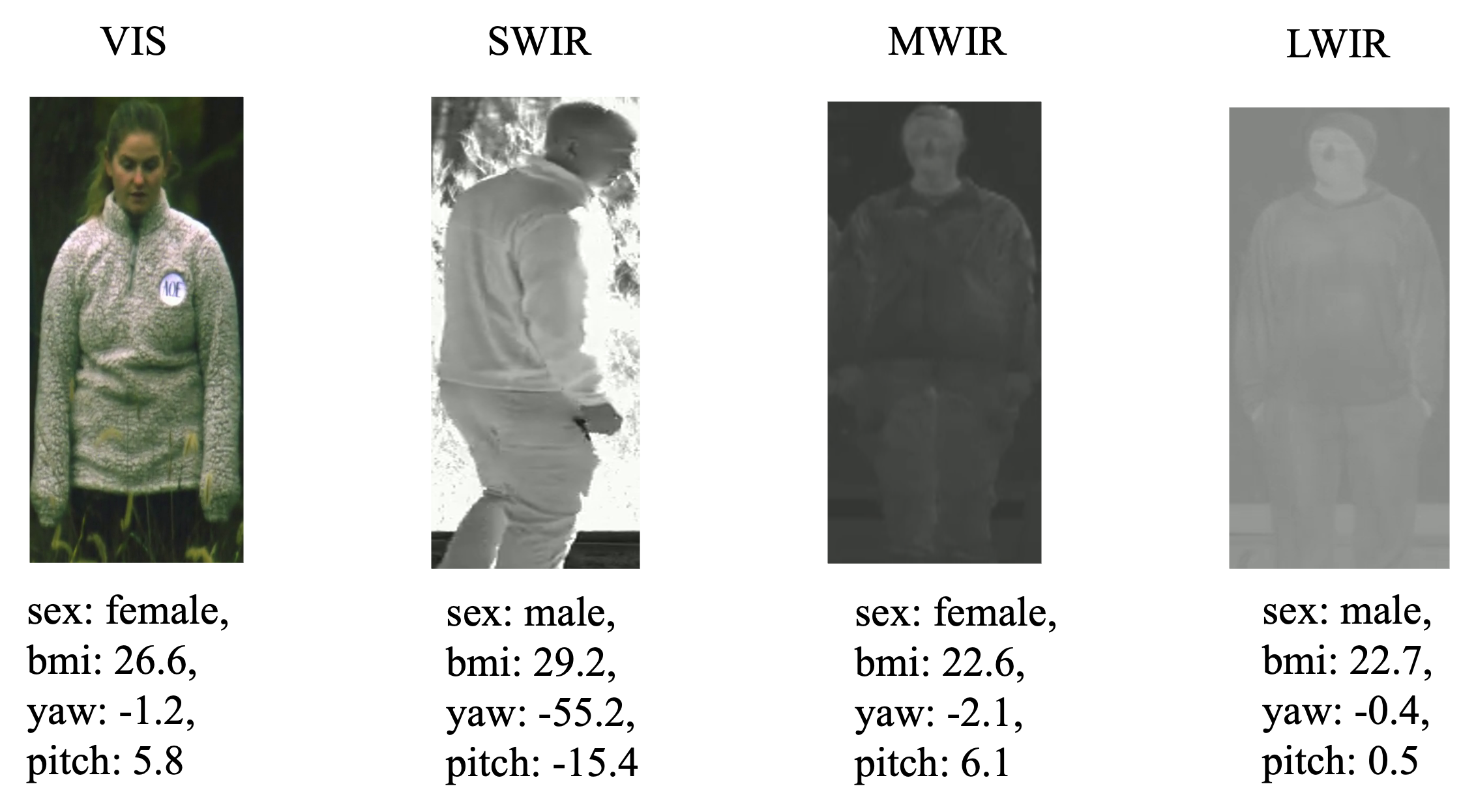}
  \caption{Attribute annotated exemplar images from the IJB-MDF dataset across visible and infrared domains. Substantial appearance differences across VIS, SWIR, MWIR, and LWIR modalities illustrate the cross-spectral challenges.}
  \label{examplesubjmdf}
\end{figure}

To analyze attribute expressivity under cross-spectral domain shifts, we additionally evaluate on the IJB-MDF dataset. This dataset contains videos of 251 subjects acquired across multiple sensing domains, including visible (VIS), short-wave infrared (SWIR), mid-wave infrared (MWIR), and long-wave infrared (LWIR), as well as long-range surveillance conditions. Following prior IARPA Janus protocols~\cite{kalka2019iarpa,nanduri2024template}, all media corresponding to a subject are treated as a template and 1:N identification is performed.


\noindent\textbf{Ground-truth estimation:}  
We use VIS-500m videos as the gallery domain and SWIR-30m, MWIR-30m, and LWIR-30m as query domains. The dataset provides ground-truth face bounding boxes. Body bounding boxes for model training are generated using YOLOv10~\cite{wang2024yolov10}, with identity labels assigned via maximum IoU overlap between body and ground-truth face bounding boxes, and ambiguous detections with IoU exceeding 0.75 across multiple faces are discarded~\cite{nanduri2025multi}. An independent audit of this legacy pipeline indicates that approximately 1.5\% of VIS-500m frames, 12.1\% of SWIR-30m frames, 8.2\% of MWIR-30m frames, and 9.5\% of LWIR-30m frames contain ambiguous identity assignments due to bounding box collisions in crowded scenes. 

Pose attributes (pitch, yaw, and roll) used in the expressivity analysis are estimated separately using the PHALP~\cite{rajasegaran2022tracking} tracker with the 4DHumans model~\cite{goel2023humans}. Identity labels are assigned by matching the predicted nose keypoint (SMPL joint index 0) to the nearest ground-truth face bounding box center, with matches accepted only when the Euclidean distance is below 50 pixels and the track is verified across at least 20 frames. Initial pose extraction achieves full coverage in the visible domain (100\%) and high coverage in SWIR (96.1\%), but lower rates in MWIR (81.9\%) and LWIR (88.4\%) due to reduced joint estimation reliability in thermal imagery. To address these gaps, temporal synchronization across the infrared sensors is exploited: for MWIR and LWIR frames where pose estimation fails, we recover angles from the corresponding SWIR capture of the same subject and frame, raising coverage to above 97\% across all infrared domains. Remaining single-frame gaps are interpolated only when the frame distance between known values is at most two frames; larger gaps are left unresolved to preserve the integrity of the pose distribution. 

Unlike BRIAR, where pose annotations are provided as part of the dataset, MDF pose attributes are estimated and may therefore introduce measurement noise that attenuates expressivity estimates. Additionally, because the training-time identity labels (YOLOv10-based) and analysis-time pose labels (4DHumans-based) are derived from separate pipelines, a small fraction of frames may carry inconsistent identity assignments. 

There are 31 videos in VIS-500m and 42 videos in each infrared domain. Figure~\ref{examplesubjmdf} illustrates the appearance gap between visible and infrared modalities.

\begin{table}[t]
\centering
\resizebox{0.8\linewidth}{!}{
\begin{tabular}{l|ccc|cc}
\toprule
\textbf{Models} & \textbf{Rank 1} & \textbf{5} & \textbf{10} & \textbf{TAR @1\%} & \textbf{@10\%} \\
\midrule
Dc-former & 27.98 & 54.06 & 62.28 & 39.84 & 84.51 \\
PFD       & 32.92 & 55.65 & 75.73 & 47.97 & 71.22 \\
SemReID   & \textbf{34.84} & \textbf{56.95} & \textbf{66.33} & \textbf{54.09} & \textbf{89.03} \\
\bottomrule
\end{tabular}
}
\vspace{0.5em}
\caption{Quantitative comparison between Dc-former, PFD, and SemReID for identification performance on BRIAR.}
\label{tab:briar_protocol2}
\end{table}

\subsection{Integration of MINE with ReID Models}

For BRIAR, we analyze attribute correlations across three state-of-the-art person ReID models: Pose-guided Feature Disentangling (PFD)~\cite{wang2022pose}, Dc-former~\cite{li2023dc}, and SemReID~\cite{huang2023self}. PFD leverages pose-guided feature disentanglement to separate visible and occluded body regions. Dc-former employs multiple class tokens within a ViT-small backbone to learn diverse embedding subspaces. SemReID introduces a Local Semantic Extraction module guided by keypoints and SAM masks to capture fine-grained body semantics.

To quantify attribute encoding, we integrate the Mutual Information Neural Estimator (MINE) as an auxiliary analysis module. For each image, features are extracted from layer \(L\) to form \( \mathbf{F} \), which is concatenated with attribute vector \( \mathbf{A} \) to produce \( \mathbf{X}=[\mathbf{F}\,|\,\mathbf{A}] \). The MINE network \(T_\theta\), implemented as a two-layer MLP (512 and 128 units with ELU activations), maximizes the Donsker–Varadhan lower bound to estimate mutual information between features and attributes. Multiple runs are averaged to obtain stable expressivity estimates.

For the IJB-MDF experiments, we restrict analysis to the SemReID backbone. This choice is motivated by its superior identification performance on the BRIAR dataset (Table~\ref{tab:briar_protocol2}), where it consistently outperforms PFD and Dc-former across Rank-1 accuracy and TAR metrics. Since cross-spectral analysis primarily aims to study how attribute encoding transfers under domain shifts rather than compare backbone architectures, focusing on the strongest-performing model provides a more stable and interpretable evaluation.

Accordingly, we apply the same MINE framework to SemReID under two training settings:

\begin{itemize}
\item \textbf{Base model:} pretrained on visible-spectrum datasets without domain adaptation.
\item \textbf{Fine-tuned model:} further trained on IJB-MDF using domain-aware sampling to improve cross-spectral robustness.
\end{itemize}

We compute expressivity for age, gender, BMI, pitch, and yaw across transformer layers. This enables layer-wise analysis of how attribute information evolves within the representation hierarchy and how cross-spectral fine-tuning modifies these correlations.
\subsection{Hierarchical and Temporal Analysis of Attribute Influence}

We analyze attribute encoding both hierarchically across layers and temporally across training epochs. SemReID and PFD use ViT-base backbones with 12 layers, while Dc-former uses ViT-small with 8 layers.

\noindent\textbf{Hierarchical Analysis:}  
For SemReID and PFD, features are extracted from layers 2, 4, 6, 9, and 12. For Dc-former, we analyze layers 2, 3, 4, 6, and 8. Early layers capture low-level spatial cues, mid layers encode structural patterns, and late layers capture semantic identity information. The same layer-wise analysis is applied to the IJB-MDF experiments using the SemReID backbone.

\noindent\textbf{Temporal Analysis:}  
For BRIAR, models are trained for 11 epochs, and expressivity is evaluated at epochs 1, 3, 5, 8, and 11. Early epochs capture the emergence of attribute encoding, while later epochs reflect stabilization as the model converges. Temporal analysis is not performed for the IJB-MDF experiments. In this setting, fine-tuning converges rapidly, typically within two epochs, as a result, there are no intermediate epochs to capture meaningful variation in attribute expressivity. Consequently, we focus on comparing the base pretrained model and the final fine-tuned model, which provides a clearer assessment of cross-spectral adaptation effects.

\subsection{Implementation Details}
The Mutual Information Neural Estimation (MINE) network is instantiated as a two-layer multilayer perceptron (MLP) with hidden dimensions 512 and 128 and ELU activations, used to parameterize the statistics network $T_\theta$ (Fig.~2). The network is initialized using Xavier normal initialization and optimized using Adam with a learning rate of $10^{-3}$ and batch size 100 until convergence of Eq.~(7). The architecture is consistent across all experiments, with only the input layer adapting to the dimensionality of the concatenated feature--attribute representation. Expressivity is computed following Algorithm~\ref{alg:expressivity}, and for each layer--attribute pair, MINE training is repeated $M=5$ times with results averaged to ensure stability.

For SemReID, we adopt Vision Transformer (ViT) backbones~\cite{dosovitskiy2020image} with input resolution $384 \times 128$. A dual-stream design extracts both global and local semantic features, where the global descriptor is 768-dimensional and local features correspond to three semantic regions (face, upper body, and lower body), each of dimension 768. Local features are averaged to form a single 768-dimensional embedding, which is concatenated with the global descriptor to yield a final 1536-dimensional representation. Multi-crop augmentation~\cite{caron2020unsupervised, caron2021emerging} is employed using $M=2$ global views and $N=3$ local views, followed by $L=12$ cross-attention layers. Identity embeddings are computed via a batch normalization layer for efficiency. The resulting feature vectors are concatenated with attribute representations and passed to the MINE network for mutual information estimation.

Dc-former~\cite{li2023dc} constructs diverse embedding subspaces by leveraging multiple class tokens, combined using a self-diverse constraint (SDC) with $\lambda = 1$. A dynamic weight controller balances the contribution of each token to form a compact and discriminative descriptor.

PFD~\cite{wang2022pose} employs overlapping patch embeddings (step size = 12) and pose-guided feature aggregation using keypoints estimated via HRNet, with a confidence threshold $\gamma = 0.2$. A part-view–based decoder with $N_v = 17$ learnable semantic views is used, with the decoder consisting of 6 layers.

For IJB-MDF, SemReID is pretrained on LUPerson~\cite{fu2021unsupervised} and subsequently fine-tuned using domain-aware sampling across visible (VIS) and infrared (SWIR, MWIR, LWIR) domains. Training uses resized frames of $384 \times 128$. Expressivity is computed for both base and fine-tuned models across layers 2, 4, 6, 9, and 12 using identical MINE hyperparameters, enabling consistent comparison across domains and training stages.

\section{Results and Discussions}

For the person re-identification task on BRIAR, SemReID achieves the strongest identification performance among the evaluated models, as shown in Table~\ref{tab:briar_protocol2}. Building on this quantitative comparison, we next examine feature--attribute correlations using the proposed MINE-based expressivity analysis. To present these findings in a structured manner, we organize the discussion into three parts: first, the progression of attribute expressivity across the hierarchical feedforward pipeline, second, the temporal evolution of expressivity during training, and third, the practical advantages of using expressivity as an analysis tool in person ReID. In addition, we extend the layer-wise analysis to a cross-domain setting to assess how attribute encoding behaves under distribution shifts. Focusing on auxiliary attributes such as BMI, pose, and demographic factors, we observe that these attributes exhibit non-trivial and structured expressivity patterns across layers. Despite not being explicitly supervised, they are consistently encoded within the learned feature representations. This indicates that person ReID models capture a mixture of attribute-related signals alongside identity-discriminative information, rather than learning fully disentangled representations. Notably, these trends persist in the cross-domain analysis, where we examine attribute expressivity across network depth. As we show in subsequent sections, analyzing these attribute-specific expressivity patterns helps us to understand how different factors emerge, evolve, and interact across network depth and during training, as well as how they transfer across domains. This perspective not only complements standard performance evaluation but also offers deeper insight into the internal structure and generalization behavior of learned representations in person ReID models.
\subsection{Hierarchical Feedforward Progression of Attribute Expressivity}
\label{sec:hierarchical_analysis}

\begin{figure}[htbp]
  \centering
  \includegraphics[width=\linewidth]{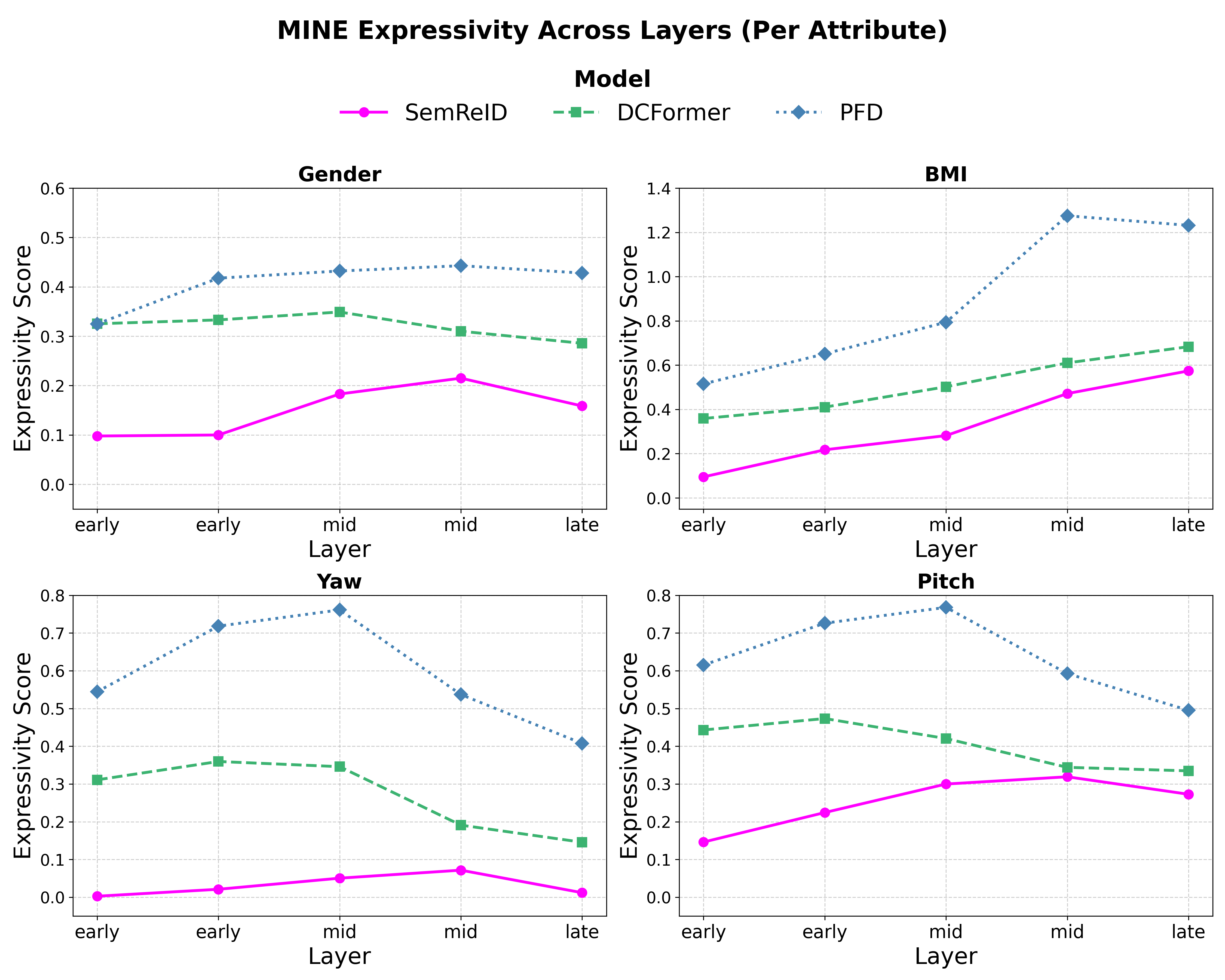}
  \caption{Expressivity trends of gender, yaw, pitch and BMI in input image over layer-wise learnt features from SemReID.}
  \vspace{-1mm}
  \label{layerfig}
\end{figure}

To investigate how attribute-related information evolves with network depth, we extract features from multiple transformer layers for each model. Specifically, for PFD and SemReID, which are based on ViT-Base, we analyze layers 2, 4, 6, 9, and 12. For DCFormer, which employs a ViT-Small backbone, we analyze layers 2, 3, 4, 6, and 8. For interpretability, we group these layers into hierarchical stages. In ViT-Base, layers 2 and 4 are regarded as \textit{early}, layers 6 and 9 as \textit{mid-level}, and layer 12 as \textit{late}. In ViT-Small, layers 2 and 3 are treated as \textit{early}, layers 4 and 6 as \textit{mid-level}, and layer 8 as \textit{late}. Figure~\ref{layerfig} illustrates the layer-wise expressivity trends.

Several consistent patterns emerge from this hierarchical analysis.

\begin{itemize}
    \item \textbf{BMI becomes increasingly prominent with depth across all models:}  
    BMI expressivity increases steadily as features propagate through the network. This trend is evident for all three models, with the most pronounced increase observed in PFD, where BMI expressivity rises from approximately 0.6 in the earlier layers to more than 1.2 in the final layer. This behavior suggests that morphometric characteristics are progressively accumulated and emphasized in deeper layers, indicating that BMI remains strongly entangled with the learned identity representation.

    \item \textbf{Yaw is strongest in intermediate layers and declines in late representations:}  
    Across all models, yaw expressivity follows a non-monotonic trend: it increases from the early layers, reaches its peak at intermediate depths, and then declines in the final block. This effect is especially pronounced in PFD, where yaw rises to approximately 0.75 in the middle of the network before dropping to roughly 0.4 in the final layer. Such a pattern suggests that pose-related orientation cues are initially useful for representation building, but are subsequently attenuated as the model shifts toward more identity-specific features.

    \item \textbf{Pitch exhibits a similar but less pronounced behavior than yaw:}  
    Pitch also tends to peak in the middle layers before decreasing in deeper layers. However, compared to yaw, the decline is less abrupt, and pitch remains moderately encoded even in later representations, particularly for DCFormer and PFD. This indicates that although pitch is partially suppressed as the hierarchy deepens, it is not discarded as aggressively as yaw and continues to contribute to the learned embedding.

    \item \textbf{Gender remains moderately encoded and relatively stable across depth:}  
    Gender expressivity shows a more stable profile than BMI or pose-related attributes. In PFD and DCFormer, gender remains relatively consistent across layers, without sharp peaks or collapses. In SemReID, it shows a modest rise in the mid-level layers followed by a decrease toward the final block. Overall, gender is less sensitive to depth than BMI, pitch, or yaw, yet it remains a persistent latent factor throughout the hierarchy.

    \item \textbf{Distinct model-specific trends are observed:}  
    Among the three models, PFD consistently exhibits the highest expressivity values across nearly all attributes and layers, indicating stronger attribute leakage and comparatively weaker suppression of non-identity information. DCFormer shows intermediate expressivity levels and preserves clear hierarchical progression. SemReID, in contrast, generally yields the lowest expressivity values, particularly for yaw and gender, suggesting that it learns a more identity-focused representation with reduced sensitivity to auxiliary attributes.
\end{itemize}

Taken together, these results show that the feedforward hierarchy does not treat all attributes uniformly. Pose-related cues such as yaw and pitch are most salient in intermediate layers, whereas BMI becomes progressively more prominent in deeper blocks. In the final-layer representations of all models, we observe a consistent ordering of attribute expressivity:
\[
\text{BMI} > \text{pitch} > \text{gender} > \text{yaw}.
\]
This ordering indicates that BMI is the most persistent attribute in the learned feature space, while yaw is the most effectively suppressed by the time the final representation is formed.

\subsection{Temporal Progression of Attribute Expressivity}
\label{sec:temporal_analysis}

\begin{figure}[htbp]
  \centering
  \includegraphics[width=\linewidth]{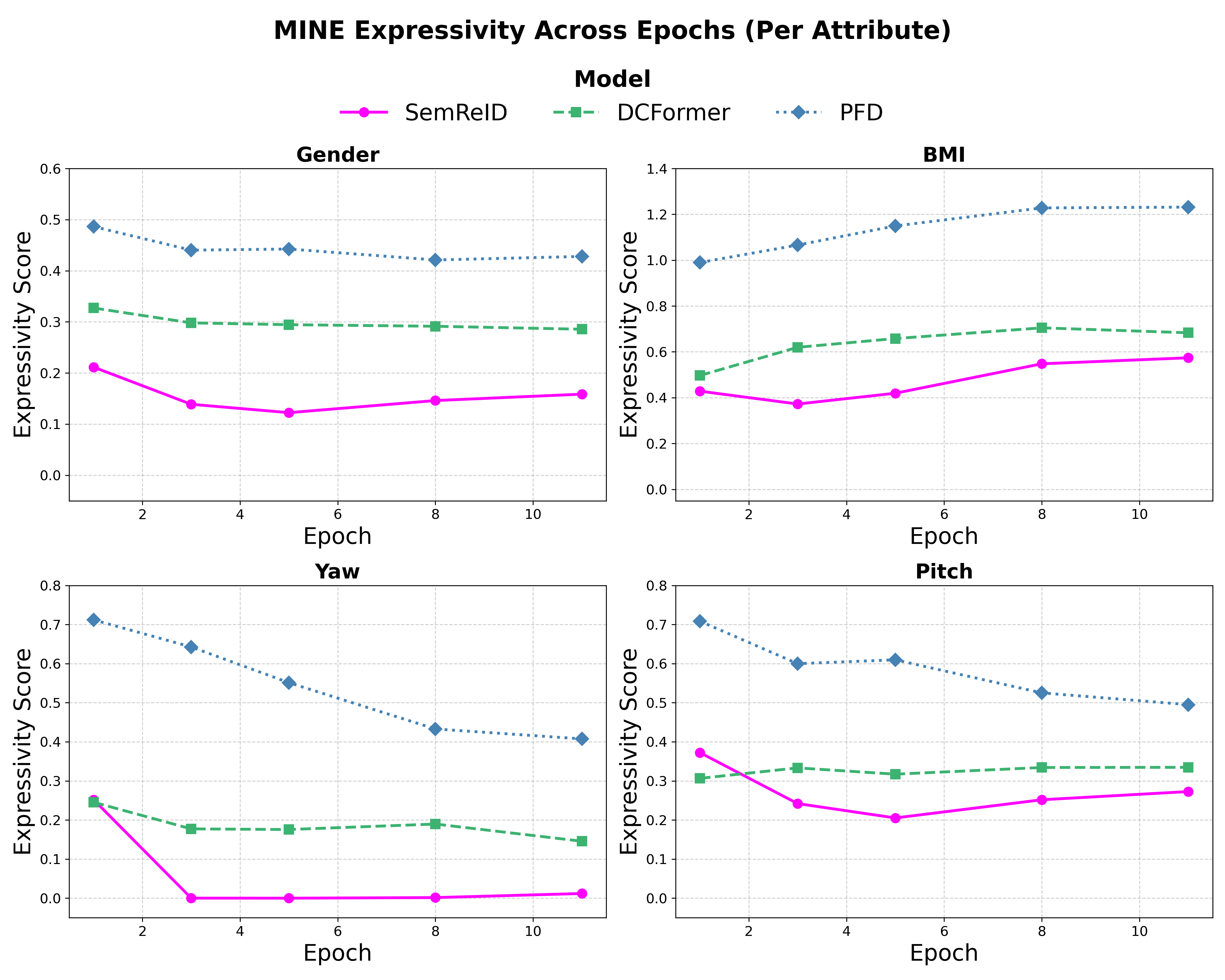}
  \caption{Expressivity trends of gender, yaw, pitch and BMI in input image over epoch-wise learnt features from SemReID.}
  \vspace{-3mm}
  \label{epochfig}
\end{figure}

We next analyze how attribute expressivity evolves over the course of training. For this temporal analysis, we evaluate MINE expressivity scores at the final attention layer of SemReID, DCFormer, and PFD across multiple epochs. Figure~\ref{epochfig} presents the epoch-wise trends. This analysis reveals how different attributes emerge, stabilize, or are suppressed as optimization progresses.

Several key observations can be drawn from these temporal trends.

\begin{itemize}
    \item \textbf{The final ranking of attribute expressivity is consistent across all models:}  
    By epoch 11, all three models converge to the same ordering of attribute expressivity:
    \[
    \text{BMI} > \text{pitch} > \text{gender} > \text{yaw}.
    \]
    This consistency suggests that BMI is the most persistently encoded attribute throughout training, whereas yaw becomes the least represented in the final feature space.

    \item \textbf{Yaw is progressively suppressed during training:}  
    All models show a substantial decline in yaw expressivity as training proceeds. In PFD, yaw decreases from above 0.7 to roughly 0.4 by epoch 11, while in SemReID it drops to nearly zero shortly after epoch 2. This sharp reduction indicates that yaw is actively suppressed as the models learn representations that are more invariant to pose and more discriminative for identity.

    \item \textbf{BMI remains persistent throughout optimization:}  
    In contrast to yaw, BMI does not diminish during training. Instead, it either remains stable or increases. PFD shows the clearest increase, with BMI expressivity rising from approximately 1.0 to 1.2. DCFormer and SemReID exhibit smaller but still consistent gains. This behavior highlights BMI as a persistent and difficult-to-remove source of entanglement in identity features, suggesting that body morphology may act as a confounding factor in person ReID.

    \item \textbf{Gender remains comparatively low and stable:}  
    Gender expressivity stays relatively low throughout training, apart from yaw which is even more strongly suppressed by the end. In SemReID, gender decreases early and stabilizes at approximately 0.13, indicating gradual decoupling from the learned identity representation. The same general trend is observed in the other models, though with somewhat higher retained values.

    \item \textbf{Pitch occupies an intermediate regime:}  
    Pitch shows behavior between BMI and yaw. It generally stabilizes after epoch 3, with DCFormer consistently retaining more pitch information than SemReID. This suggests that pitch is neither fully suppressed nor strongly amplified during training, but instead remains a moderately persistent factor in the final representation.

    \item \textbf{The models differ substantially in how much attribute information they retain:}  
    PFD consistently encodes the largest amount of attribute information across nearly all attributes and epochs. SemReID, on the other hand, exhibits the lowest overall expressivity, indicating that it learns more robust identity representations with reduced dependence on spurious cues such as yaw and gender. DCFormer again occupies an intermediate position between the two.
\end{itemize}

Overall, the temporal analysis reveals that training does not affect all attributes equally. Pose-related information, especially yaw, is progressively removed as the models optimize for identity discrimination, whereas morphometric information such as BMI remains embedded in the learned feature space. These findings suggest that although existing architectures can naturally suppress some auxiliary factors, persistent attributes such as BMI may require more targeted disentanglement or regularization strategies.

\subsection{Advantages of Expressivity for Person ReID}

In this subsection, we further motivate the use of MINE-based expressivity as an analysis tool relative to existing alternatives. Two practical advantages are particularly important in the person ReID setting:

\begin{enumerate}
    \item \textbf{It supports both discrete and continuous attributes:}  
    Expressivity provides a unified framework that can be applied to both discrete attributes, such as gender, and continuous attributes, such as pitch angle or BMI. For example, gender can be represented using a binary attribute vector \(A\), whereas pose and BMI can be modeled directly as continuous variables without discretization. This flexibility is a key advantage over methods such as TCAV~\cite{kim2018interpretability}, which are more naturally suited to discrete concept definitions and rely on the availability of well-defined negative examples. In person ReID, where many relevant factors are continuous rather than categorical, expressivity therefore offers a more natural and general analysis framework.

    \item \textbf{It is independent of training identity classes:}  
    Several prior analysis methods rely on changes in logits or class-specific outputs, which restricts them to identities seen during training. In contrast, expressivity does not depend on logit differences or the presence of training identity labels in the analysis stage. Instead, it directly measures the statistical dependency between learned features and attribute values. This makes it particularly useful for studying unseen identities and attributes that are not explicitly modeled during training, which is highly relevant in unconstrained person ReID settings.
\end{enumerate}
\subsection{Expressivity Analysis on the MDF Dataset}
\label{sec:mdf_analysis}

To evaluate how feature--attribute correlations behave under sensing-domain shift, we extend the expressivity analysis of the SemReID backbone to the IJB-MDF dataset. Unlike the BRIAR experiments, which compare multiple architectures, MDF evaluates a single backbone under two training settings: (1) a base model trained only on visible-spectrum images, and (2) the same model after cross-spectral fine-tuning across Visible, SWIR, MWIR, and LWIR domains. We compute MINE expressivity for five attributes: BMI, gender, age, yaw, and pitch. In both settings, we probe representative \textit{early}, \textit{mid}, and \textit{late} layers (with two checkpoints each for early and mid stages) to track how attribute information accumulates through network depth.

The MDF experiments introduce two complementary levels of domain shift. First, there is a dataset-level gap between the pretraining data (LUPerson) and the target MDF dataset, involving differences in scene composition, capture conditions, and identity distribution. Second, within MDF, there exists a spectral domain gap between visible imagery and infrared modalities (SWIR, MWIR, and LWIR), which alters the underlying sensing physics and available visual cues. Evaluating expressivity under these two levels of shift allows us to study the robustness of attribute encoding both across datasets and across sensing modalities.

To provide a reference scale for attribute expressivity, we additionally compute expressivity for identity itself. Specifically, subject identity is treated as a categorical attribute by assigning each identity a unique scalar value, allowing it to be incorporated into the same MINE-based estimation framework used for other attributes. This formulation enables direct comparison between identity-related information and auxiliary attributes within the learned representation. As expected, identity expressivity is consistently higher than all other attributes and increases monotonically with network depth. Quantitatively, identity expressivity typically lies in the range of 4–6 across layers, whereas the strongest auxiliary attributes (e.g., BMI and pitch) saturate around 2–2.5, indicating that identity-related information is encoded at approximately twice the magnitude of auxiliary factors. This separation suggests that deeper layers progressively concentrate identity-discriminative information while maintaining controlled encoding of secondary attributes. The consistent growth of identity expressivity also provides a sanity check for the estimator, confirming that the learned representations increasingly emphasize identity-related structure, which is the primary objective of person re-identification models.

\paragraph{Base Model Trained on Visible Spectrum.}

\begin{figure*}[t]
\centering
\includegraphics[width=\linewidth]{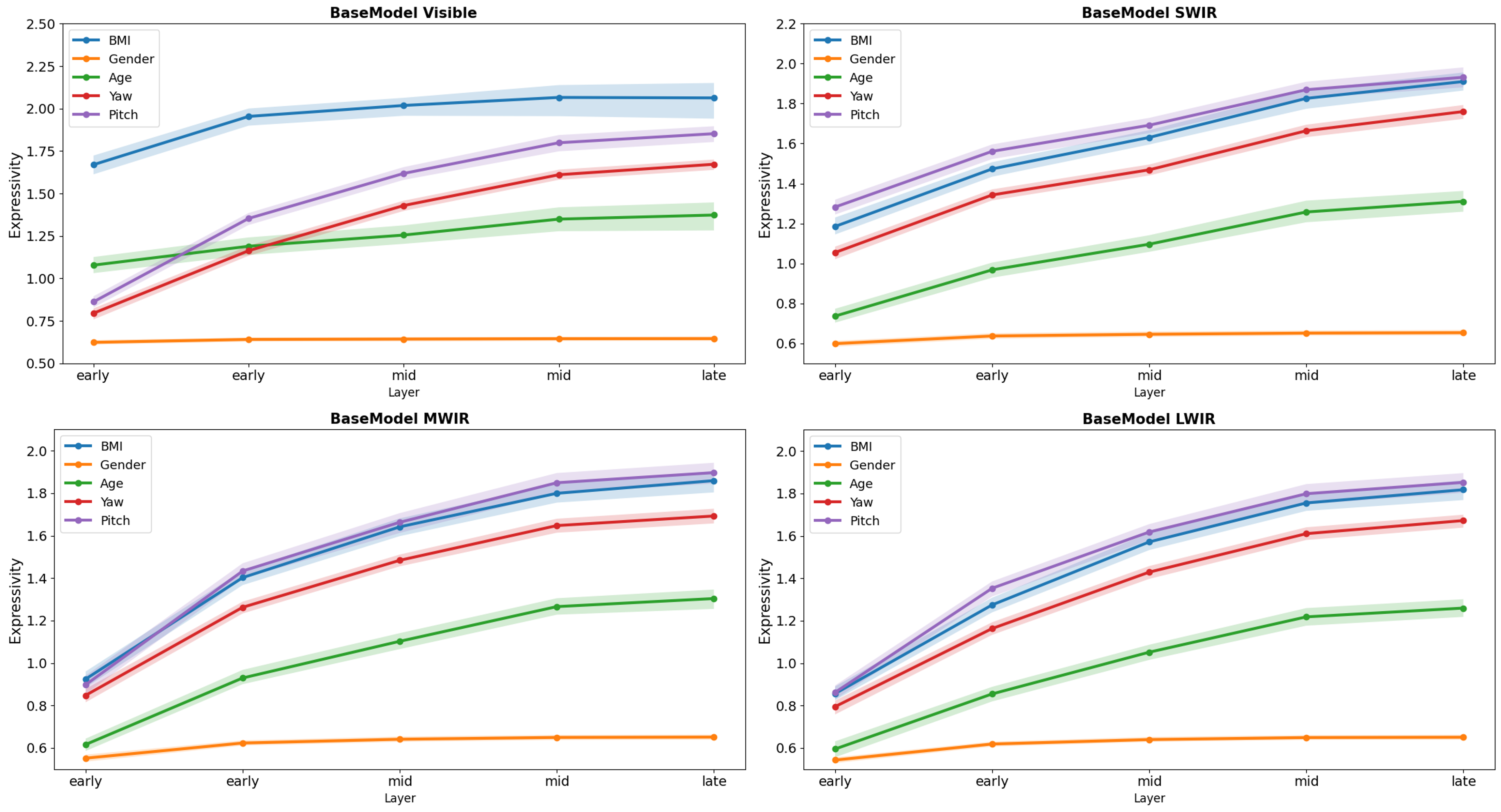}
\caption{Layer-wise attribute expressivity on the MDF dataset for the base model trained only on the visible spectrum. Expressivity scores are computed using MINE for BMI, gender, age, yaw, and pitch across hierarchical network layers for four sensing modalities: Visible, SWIR, MWIR, and LWIR. Shaded regions denote confidence intervals across runs.}
\label{fig:mdf_base}
\end{figure*}
\begin{figure*}[t]
\centering
\includegraphics[width=\linewidth]{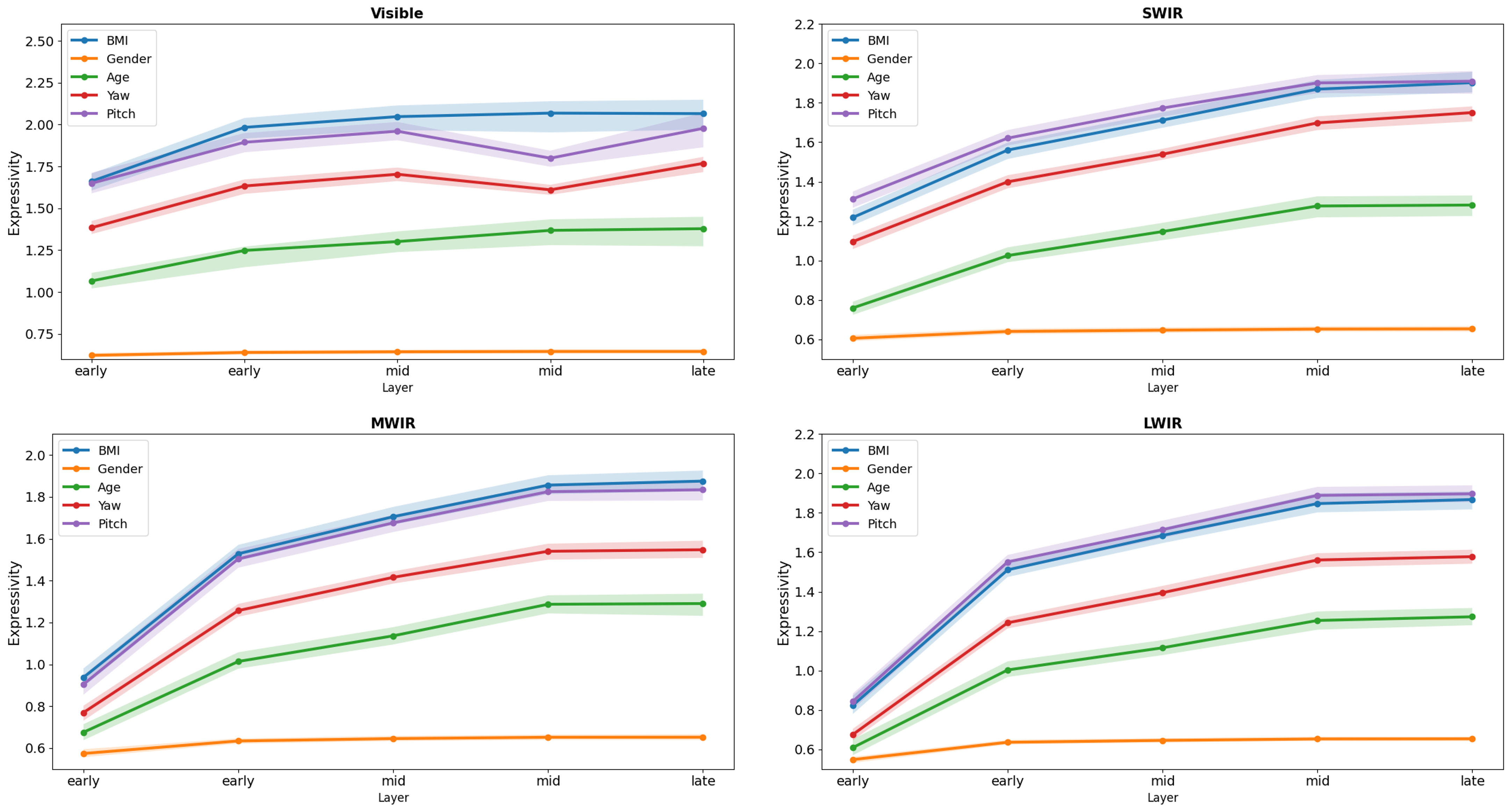}
\caption{Layer-wise attribute expressivity on the MDF dataset after cross-spectral fine-tuning. The model is adapted across Visible, SWIR, MWIR, and LWIR modalities. MINE scores illustrate how correlations between learned features and attributes evolve across hierarchical layers after multi-spectral training. Shaded regions denote confidence intervals across runs.}
\label{fig:mdf_finetuned}
\end{figure*}

Figure~\ref{fig:mdf_base} reports the layer-wise expressivity trends when the model is trained solely on visible imagery but evaluated across multiple sensing modalities. Three global observations emerge.

\textbf{(i) Hierarchical amplification across depth.}  
Across all modalities, expressivity for BMI, pitch, yaw, and age increases monotonically with network depth. Early layers capture weak correlations, while mid and late layers progressively amplify attribute information. This hierarchical growth suggests that high-level semantic representations accumulate structural and geometric cues correlated with these attributes even when the input sensing modality differs from the training domain.

The largely monotonic increase observed across attributes can be further understood by examining the attribute distributions in MDF. Unlike BRIAR, which exhibits substantial variability in pose, body composition, and demographic factors, the MDF dataset is characterized by relatively constrained attribute ranges. Age and BMI distributions are concentrated within narrow intervals, gender remains a simple binary attribute with moderate imbalance, and both pitch and yaw are strongly centered around near-frontal viewpoints across all sensing modalities. This reduced attribute variability limits intra-class diversity and diminishes competition between identity-related and auxiliary factors during representation formation. Consequently, deeper layers progressively accumulate discriminative information in a more uniform manner, leading to monotonic increases in expressivity across attributes. This behavior suggests that when attribute variation is limited, hierarchical representations tend to encode auxiliary cues consistently alongside identity, rather than selectively emphasizing or suppressing specific factors.

\textbf{(ii) Morphometrics and pose dominate identity-related structure.}  
Across modalities, BMI and pose (pitch and yaw) consistently emerge as the strongest signals. Age follows with moderate expressivity, while gender remains consistently low. This produces a stable ordering:
\[
\text{BMI} \approx \text{Pitch} \;>\; \text{Yaw} \;>\; \text{Age} \;\gg\; \text{Gender}.
\]
The persistence of this ordering across sensing domains suggests that morphometric structure and head pose are among the most robust cues captured by identity embeddings.

\textbf{(iii) Gender remains weak and depth-insensitive.}  
Gender shows minimal growth across depth and maintains low expressivity in all modalities. This indicates that gender-related cues are either weakly correlated with identity representations or not strongly amplified by the SemReID backbone.

Beyond these global trends, the infrared modalities exhibit distinct behaviors that reveal how sensing physics interacts with learned representations.

\textbf{SWIR (Short-Wave Infrared).}  
SWIR displays a smoother and more monotonic expressivity trajectory across depth. While early-layer signals are weaker than in MWIR, the gradual increase produces strong late-layer correlations for BMI and pitch. Compared to MWIR, SWIR maintains slightly more balanced growth across attributes, suggesting that SWIR imagery preserves both morphological and geometric information but with less pronounced early-layer amplification.

\textbf{MWIR (Mid-Wave Infrared).}  
MWIR exhibits the most rapid early-layer growth among the infrared domains. From early to mid layers, BMI and pitch increase sharply, approaching the magnitudes observed in the visible spectrum. This suggests that MWIR retains substantial geometric and thermal structure related to body morphology and head orientation. By late layers, pitch and BMI become nearly comparable in magnitude, indicating that pose cues remain strongly embedded in the representation even when the model was trained exclusively on RGB imagery.

\textbf{LWIR (Long-Wave Infrared).}  
LWIR begins with the lowest early-layer expressivity among the infrared modalities but demonstrates steady growth across depth. Because LWIR primarily captures thermal emission rather than reflected light, fine-grained texture cues are reduced. As a result, early layers contain limited attribute information, but deeper layers still recover strong correlations with BMI and pose, indicating that structural silhouette and thermal contours provide sufficient cues for these attributes.

\textbf{Visible Spectrum Baseline.}  
In the visible domain, BMI is consistently the most expressive attribute across all layers. This reflects the availability of rich appearance cues such as clothing geometry, body proportions, and shading patterns. Pitch and yaw follow closely, reinforcing the observation that pose remains a strongly encoded component of identity representations.

Overall, the base model demonstrates that attribute correlations remain surprisingly stable across sensing domains, even without explicit cross-spectral training. The persistence of pose and morphometric signals suggests that identity representations rely heavily on structural cues that are largely modality-agnostic. These trends are further notable given the measurement noise and identity label ambiguity documented in Section IV-A.

\paragraph{Cross-Spectral Fine-Tuned Model.}

Figure~\ref{fig:mdf_finetuned} presents the same analysis after cross-spectral fine-tuning. While the overall attribute hierarchy remains largely unchanged, two important refinements emerge.

\textbf{(i) Cross-modal trajectories become more aligned.}  
After adaptation, the growth curves for SWIR, MWIR, and LWIR more closely resemble the visible-domain trajectories. Early-layer signals increase slightly, and mid-layer transitions become smoother across modalities. This indicates that cross-spectral training encourages the backbone to learn representations that are more consistent across sensing domains.

\textbf{(ii) Pose correlations become slightly moderated.}  
While pitch and yaw remain strong signals, their growth becomes more stable after fine-tuning. In certain modalities (notably visible and MWIR), yaw shows slight attenuation in late layers compared to the base model. This suggests that cross-spectral adaptation may partially reduce over-reliance on pose while maintaining strong identity cues.

Modality-specific effects after adaptation. After fine-tuning, MWIR trajectories closely mirror those of the visible domain, indicating improved alignment between thermal and RGB representations. SWIR maintains its smooth hierarchical growth but with slightly higher early-layer expressivity, suggesting improved integration of SWIR features into the shared representation space. LWIR shows the largest relative improvement after fine-tuning, particularly in early layers, indicating that cross-spectral adaptation helps recover attribute cues that were weak in the base model.

\paragraph{Implications for Identity Representation.}

Across both training settings, the overall attribute ordering remains stable:
\[
\text{BMI} \approx \text{Pitch} \;>\; \text{Yaw} \;>\; \text{Age} \;\gg\; \text{Gender}.
\]

Importantly, the strong BMI signal should not automatically be interpreted as undesirable bias. Instead, it indicates that morphometric structure captured by BMI contributes meaningful information to identity representations. Physical body morphology is inherently linked to identity cues such as body shape, skeletal proportions, and posture. The goal is therefore not to eliminate BMI or pose information entirely, but rather to ensure that recognition performance is not overly dependent on these attributes.

Consequently, a desirable representation should balance two objectives: preserving the discriminative value of morphometric cues while preventing recognition accuracy from becoming disproportionately contingent on them. This observation motivates future work on representation disentanglement and attribute-aware regularization strategies that maintain identity performance while mitigating unintended biases.

\section{Conclusion}

We introduced an information-theoretic framework to quantify how much attribute-relevant information transformer-based person re-identification networks encode without explicit supervision. By estimating expressivity using Mutual Information Neural Estimation (MINE), our approach provides a direct measure of feature--attribute dependence and enables systematic analysis across hierarchical layers, training dynamics, and sensing modalities.

Across visible-spectrum person ReID on the BRIAR dataset, several consistent findings emerge. BMI exhibits the highest expressivity and increases with both network depth and training progression, indicating that morphometric cues are persistently embedded in identity representations even without explicit supervision. Pose attributes such as yaw and pitch are strongly encoded in intermediate layers but are attenuated in deeper layers as representations become more identity-focused, with yaw showing the strongest suppression during training. Gender remains moderately encoded with minimal variation across the feature hierarchy. These trends suggest that identity representations evolve from capturing general geometric factors toward more stable morphometric structure as training progresses.

Extending the analysis to cross-spectral identification on the MDF dataset (VIS, SWIR, MWIR, LWIR) demonstrates that the dominance of BMI persists across sensing modalities, underscoring morphology as a robust identity cue under modality shifts. Age shows consistent growth across layers, suggesting stable retention of coarse physiological traits even when sensing physics changes. In contrast to BRIAR, pose attributes exhibit more monotonic growth across depth in MDF, which is consistent with the reduced pose variability and narrower attribute distributions within this dataset. Gender remains the weakest signal across spectra, indicating limited contribution to identity representation.

To contextualize attribute magnitudes, we additionally evaluated identity expressivity as a reference. Identity-related information increases consistently with network depth and exceeds auxiliary attributes, confirming that deeper layers progressively concentrate identity-discriminative structure. This observation supports the interpretation that auxiliary attributes are encoded alongside, but remain secondary to, the primary identity objective.

Overall, the results demonstrate that modern transformer-based ReID embeddings encode a structured hierarchy of implicit attributes whose influence varies across depth, training, and sensing domains. Morphometric cues, particularly BMI, emerge as persistent identity-related signals, while pose contributes transiently during representation formation and gender contributes minimally. These trends remain stable under both dataset-level and spectral domain shifts, indicating that identity representations rely heavily on modality-agnostic structural cues.

Finally, because expressivity approximates mutual information, absolute values depend on attribute entropy and label distribution. Consequently, cross-attribute comparisons should be interpreted cautiously, consistent with known limitations of MI-based analyses. Despite this limitation, the proposed framework provides a principled and generalizable tool for analyzing implicit attribute encoding in deep representation learning and offers insights for designing more robust and bias-aware person re-identification systems.


%





\ifCLASSOPTIONcaptionsoff
  \newpage
\fi



%

{\small
\bibliographystyle{IEEEtran}
\bibliography{Transactions-Bibliography/IEEEabrv.bib}

@article{hill2019deep,
  title={Deep convolutional neural networks in the face of caricature},
  author={Hill, Matthew Q and Parde, Connor J and Castillo, Carlos D and Colon, Y Ivette and Ranjan, Rajeev and Chen, Jun-Cheng and Blanz, Volker and O’Toole, Alice J},
  journal={Nature Machine Intelligence},
  volume={1},
  number={11},
  pages={522--529},
  year={2019},
  publisher={Nature Publishing Group UK London}
}

@article{nagpal2019deep,
  title={Deep learning for face recognition: Pride or prejudiced?},
  author={Nagpal, Shruti and Singh, Maneet and Singh, Richa and Vatsa, Mayank},
  journal={arXiv preprint arXiv:1904.01219},
  year={2019}
}

@inproceedings{parde2017face,
  title={Face and image representation in deep CNN features},
  author={Parde, Connor J and Castillo, Carlos and Hill, Matthew Q and Colon, Y Ivette and Sankaranarayanan, Swami and Chen, Jun-Cheng and O’Toole, Alice J},
  booktitle={2017 12th ieee international conference on automatic face \& gesture recognition (fg 2017)},
  pages={673--680},
  year={2017},
  organization={IEEE}
}

@article{givens2013introduction,
  title={Introduction to face recognition and evaluation of algorithm performance},
  author={Givens, Geof H and Beveridge, J Ross and Phillips, P Jonathon and Draper, Bruce and Lui, Yui Man and Bolme, David},
  journal={Computational Statistics \& Data Analysis},
  volume={67},
  pages={236--247},
  year={2013},
  publisher={Elsevier}
}

@inproceedings{lee2014generalizing,
  title={Generalizing face quality and factor measures to video},
  author={Lee, Yooyoung and Phillips, P Jonathon and Filliben, James J and Beveridge, J Ross and Zhang, Hao},
  booktitle={IEEE International Joint Conference on Biometrics},
  pages={1--8},
  year={2014},
  organization={IEEE}
}

@inproceedings{dhar2020attributes,
  title={How are attributes expressed in face DCNNs?},
  author={Dhar, Prithviraj and Bansal, Ankan and Castillo, Carlos D and Gleason, Joshua and Phillips, P Jonathon and Chellappa, Rama},
  booktitle={2020 15th IEEE International Conference on Automatic Face and Gesture Recognition (FG 2020)},
  pages={85--92},
  year={2020},
  organization={IEEE}
}

@article{behera2020person,
  title={Person re-identification for smart cities: State-of-the-art and the path ahead},
  author={Behera, Nayan Kumar Subhashis and Sa, Pankaj Kumar and Bakshi, Sambit},
  journal={Pattern Recognition Letters},
  volume={138},
  pages={282--289},
  year={2020},
  publisher={Elsevier}
}

@article{khan2024deep,
  title={Deep-ReID: Deep features and autoencoder assisted image patching strategy for person re-identification in smart cities surveillance},
  author={Khan, Samee Ullah and Hussain, Tanveer and Ullah, Amin and Baik, Sung Wook},
  journal={Multimedia Tools and Applications},
  volume={83},
  number={5},
  pages={15079--15100},
  year={2024},
  publisher={Springer}
}

@article{camara2020pedestrian,
  title={Pedestrian models for autonomous driving part ii: high-level models of human behavior},
  author={Camara, Fanta and Bellotto, Nicola and Cosar, Serhan and Weber, Florian and Nathanael, Dimitris and Althoff, Matthias and Wu, Jingyuan and Ruenz, Johannes and Dietrich, Andr{\'e} and Markkula, Gustav and others},
  journal={IEEE Transactions on Intelligent Transportation Systems},
  volume={22},
  number={9},
  pages={5453--5472},
  year={2020},
  publisher={IEEE}
}

@inproceedings{wong2020identifying,
  title={Identifying unknown instances for autonomous driving},
  author={Wong, Kelvin and Wang, Shenlong and Ren, Mengye and Liang, Ming and Urtasun, Raquel},
  booktitle={Conference on Robot Learning},
  pages={384--393},
  year={2020},
  organization={PMLR}
}

@inproceedings{gu2022clothes,
  title={Clothes-changing person re-identification with rgb modality only},
  author={Gu, Xinqian and Chang, Hong and Ma, Bingpeng and Bai, Shutao and Shan, Shiguang and Chen, Xilin},
  booktitle={Proceedings of the IEEE/CVF conference on computer vision and pattern recognition},
  pages={1060--1069},
  year={2022}
}

@inproceedings{gu2019temporal,
  title={Temporal knowledge propagation for image-to-video person re-identification},
  author={Gu, Xinqian and Ma, Bingpeng and Chang, Hong and Shan, Shiguang and Chen, Xilin},
  booktitle={Proceedings of the IEEE/CVF international conference on computer vision},
  pages={9647--9656},
  year={2019}
}

@inproceedings{huang2019celebrities,
  title={Celebrities-reid: A benchmark for clothes variation in long-term person re-identification},
  author={Huang, Yan and Wu, Qiang and Xu, Jingsong and Zhong, Yi},
  booktitle={2019 International Joint Conference on Neural Networks (IJCNN)},
  pages={1--8},
  year={2019},
  organization={IEEE}
}

@inproceedings{zheng2015scalable,
  title={Scalable person re-identification: A benchmark},
  author={Zheng, Liang and Shen, Liyue and Tian, Lu and Wang, Shengjin and Wang, Jingdong and Tian, Qi},
  booktitle={Proceedings of the IEEE international conference on computer vision},
  pages={1116--1124},
  year={2015}
}

@inproceedings{cao2023event,
  title={Event-guided person re-identification via sparse-dense complementary learning},
  author={Cao, Chengzhi and Fu, Xueyang and Liu, Hongjian and Huang, Yukun and Wang, Kunyu and Luo, Jiebo and Zha, Zheng-Jun},
  booktitle={Proceedings of the IEEE/CVF Conference on Computer Vision and Pattern Recognition},
  pages={17990--17999},
  year={2023}
}

@inproceedings{hou2020temporal,
  title={Temporal complementary learning for video person re-identification},
  author={Hou, Ruibing and Chang, Hong and Ma, Bingpeng and Shan, Shiguang and Chen, Xilin},
  booktitle={Computer Vision--ECCV 2020: 16th European Conference, Glasgow, UK, August 23--28, 2020, Proceedings, Part XXV 16},
  pages={388--405},
  year={2020},
  organization={Springer}
}

@inproceedings{yan2020learning,
  title={Learning multi-granular hypergraphs for video-based person re-identification},
  author={Yan, Yichao and Qin, Jie and Chen, Jiaxin and Liu, Li and Zhu, Fan and Tai, Ying and Shao, Ling},
  booktitle={Proceedings of the IEEE/CVF conference on computer vision and pattern recognition},
  pages={2899--2908},
  year={2020}
}

@inproceedings{zhang2020multi,
  title={Multi-granularity reference-aided attentive feature aggregation for video-based person re-identification},
  author={Zhang, Zhizheng and Lan, Cuiling and Zeng, Wenjun and Chen, Zhibo},
  booktitle={Proceedings of the IEEE/CVF conference on computer vision and pattern recognition},
  pages={10407--10416},
  year={2020}
}

@inproceedings{wu2022cavit,
  title={Cavit: Contextual alignment vision transformer for video object re-identification},
  author={Wu, Jinlin and He, Lingxiao and Liu, Wu and Yang, Yang and Lei, Zhen and Mei, Tao and Li, Stan Z},
  booktitle={European Conference on Computer Vision},
  pages={549--566},
  year={2022},
  organization={Springer}
}

@inproceedings{cornett2023expanding,
  title={Expanding accurate person recognition to new altitudes and ranges: The briar dataset},
  author={Cornett, David and Brogan, Joel and Barber, Nell and Aykac, Deniz and Baird, Seth and Burchfield, Nicholas and Dukes, Carl and Duncan, Andrew and Ferrell, Regina and Goddard, Jim and others},
  booktitle={Proceedings of the IEEE/CVF Winter Conference on Applications of Computer Vision},
  pages={593--602},
  year={2023}
}

@inproceedings{liu2024farsight,
  title={Farsight: A physics-driven whole-body biometric system at large distance and altitude},
  author={Liu, Feng and Ashbaugh, Ryan and Chimitt, Nicholas and Hassan, Najmul and Hassani, Ali and Jaiswal, Ajay and Kim, Minchul and Mao, Zhiyuan and Perry, Christopher and Ren, Zhiyuan and others},
  booktitle={Proceedings of the IEEE/CVF Winter Conference on Applications of Computer Vision},
  pages={6227--6236},
  year={2024}
}

@inproceedings{nikhal2024hashreid,
  title={HashReID: Dynamic Network with Binary Codes for Efficient Person Re-identification},
  author={Nikhal, Kshitij and Ma, Yujunrong and Bhattacharyya, Shuvra S and Riggan, Benjamin S},
  booktitle={Proceedings of the IEEE/CVF Winter Conference on Applications of Computer Vision},
  pages={6046--6055},
  year={2024}
}

@inproceedings{nikhal2023weakly,
  title={Weakly supervised face and whole body recognition in turbulent environments},
  author={Nikhal, Kshitij and Riggan, Benjamin S},
  booktitle={2023 IEEE International Joint Conference on Biometrics (IJCB)},
  pages={1--10},
  year={2023},
  organization={IEEE}
}

@inproceedings{zhu2024sharc,
  title={Sharc: Shape and appearance recognition for person identification in-the-wild},
  author={Zhu, Haidong and Zheng, Wanrong and Zheng, Zhaoheng and Nevatia, Ram},
  booktitle={Proceedings of the IEEE/CVF Winter Conference on Applications of Computer Vision},
  pages={6290--6300},
  year={2024}
}

@inproceedings{chen2023beyond,
  title={Beyond appearance: a semantic controllable self-supervised learning framework for human-centric visual tasks},
  author={Chen, Weihua and Xu, Xianzhe and Jia, Jian and Luo, Hao and Wang, Yaohua and Wang, Fan and Jin, Rong and Sun, Xiuyu},
  booktitle={Proceedings of the IEEE/CVF conference on computer vision and pattern recognition},
  pages={15050--15061},
  year={2023}
}

@inproceedings{wang2018learning,
  title={Learning discriminative features with multiple granularities for person re-identification},
  author={Wang, Guanshuo and Yuan, Yufeng and Chen, Xiong and Li, Jiwei and Zhou, Xi},
  booktitle={Proceedings of the 26th ACM international conference on Multimedia},
  pages={274--282},
  year={2018}
}

@inproceedings{zhu2022pass,
  title={Pass:Part-aware self-supervised pre-training for person re-identification},
  author={Zhu, Kuan and Guo, Haiyun and Yan, Tianyi and Zhu, Yousong and Wang, Jinqiao and Tang, Ming},
  booktitle={European conference on computer vision},
  pages={198--214},
  year={2022},
  organization={Springer}
}

@article{huang2023self,
  title={Self-Supervised Learning of Whole and Component-Based Semantic Representations for Person Re-Identification},
  author={Huang, Siyuan and Zhou, Yifan and Prabhakar, Ram and Liu, Xijun and Guo, Yuxiang and Yi, Hongrui and Peng, Cheng and Chellappa, Rama and Lau, Chun Pong},
  journal={arXiv preprint arXiv:2311.17074},
  year={2023}
}

@inproceedings{schumann2017person,
  title={Person re-identification by deep learning attribute-complementary information},
  author={Schumann, Arne and Stiefelhagen, Rainer},
  booktitle={Proceedings of the IEEE conference on computer vision and pattern recognition workshops},
  pages={20--28},
  year={2017}
}

@inproceedings{myers2023recognizing,
  title={Recognizing people by body shape using deep networks of images and words},
  author={Myers, Blake A and Jaggernauth, Lucas and Metz, Thomas M and Hill, Matthew Q and Gandi, Veda Nandan and Castillo, Carlos D and O’Toole, Alice J},
  booktitle={2023 IEEE International Joint Conference on Biometrics (IJCB)},
  pages={1--8},
  year={2023},
  organization={IEEE}
}

@inproceedings{yin2019towards,
  title={Towards interpretable face recognition},
  author={Yin, Bangjie and Tran, Luan and Li, Haoxiang and Shen, Xiaohui and Liu, Xiaoming},
  booktitle={Proceedings of the IEEE/CVF International Conference on Computer Vision},
  pages={9348--9357},
  year={2019}
}

@article{kim2014bayesian,
  title={The bayesian case model: A generative approach for case-based reasoning and prototype classification},
  author={Kim, Been and Rudin, Cynthia and Shah, Julie A},
  journal={Advances in neural information processing systems},
  volume={27},
  year={2014}
}

@inproceedings{kim2018interpretability,
  title={Interpretability beyond feature attribution: Quantitative testing with concept activation vectors (tcav)},
  author={Kim, Been and Wattenberg, Martin and Gilmer, Justin and Cai, Carrie and Wexler, James and Viegas, Fernanda and others},
  booktitle={International conference on machine learning},
  pages={2668--2677},
  year={2018},
  organization={PMLR}
}

@article{alain2016understanding,
  title={Understanding intermediate layers using linear classifier probes},
  author={Alain, Guillaume},
  journal={arXiv preprint arXiv:1610.01644},
  year={2016}
}

@inproceedings{koh2017understanding,
  title={Understanding black-box predictions via influence functions},
  author={Koh, Pang Wei and Liang, Percy},
  booktitle={International conference on machine learning},
  pages={1885--1894},
  year={2017},
  organization={PMLR}
}

@inproceedings{selvaraju2017grad,
  title={Grad-cam: Visual explanations from deep networks via gradient-based localization},
  author={Selvaraju, Ramprasaath R and Cogswell, Michael and Das, Abhishek and Vedantam, Ramakrishna and Parikh, Devi and Batra, Dhruv},
  booktitle={Proceedings of the IEEE international conference on computer vision},
  pages={618--626},
  year={2017}
}

@inproceedings{chattopadhay2018grad,
  title={Grad-cam++: Generalized gradient-based visual explanations for deep convolutional networks},
  author={Chattopadhay, Aditya and Sarkar, Anirban and Howlader, Prantik and Balasubramanian, Vineeth N},
  booktitle={2018 IEEE winter conference on applications of computer vision (WACV)},
  pages={839--847},
  year={2018},
  organization={IEEE}
}

@inproceedings{chen2021explainable,
  title={Explainable person re-identification with attribute-guided metric distillation},
  author={Chen, Xiaodong and Liu, Xinchen and Liu, Wu and Zhang, Xiao-Ping and Zhang, Yongdong and Mei, Tao},
  booktitle={Proceedings of the IEEE/CVF international conference on computer vision},
  pages={11813--11822},
  year={2021}
}

@inproceedings{tishby2015deep,
  title={Deep learning and the information bottleneck principle},
  author={Tishby, Naftali and Zaslavsky, Noga},
  booktitle={2015 ieee information theory workshop (itw)},
  pages={1--5},
  year={2015},
  organization={IEEE}
}

@inproceedings{belghazi2018mutual,
  title={Mutual information neural estimation},
  author={Belghazi, Mohamed Ishmael and Baratin, Aristide and Rajeshwar, Sai and Ozair, Sherjil and Bengio, Yoshua and Courville, Aaron and Hjelm, Devon},
  booktitle={International conference on machine learning},
  pages={531--540},
  year={2018},
  organization={PMLR}
}

@article{dosovitskiy2020image,
  title={An image is worth 16x16 words},
  author={Dosovitskiy, Alexey and Beyer, Lucas and Kolesnikov, Alexander and Weissenborn, Dirk and Zhai, Xiaohua and Unterthiner, Thomas and Dehghani, Mostafa and Minderer, Matthias and Heigold, Georg and Gelly, Sylvain and others},
  journal={arXiv preprint arXiv:2010.11929},
  volume={7},
  year={2020}
}

@article{caron2020unsupervised,
  title={Unsupervised learning of visual features by contrasting cluster assignments},
  author={Caron, Mathilde and Misra, Ishan and Mairal, Julien and Goyal, Priya and Bojanowski, Piotr and Joulin, Armand},
  journal={Advances in neural information processing systems},
  volume={33},
  pages={9912--9924},
  year={2020}
}

@inproceedings{caron2021emerging,
  title={Emerging properties in self-supervised vision transformers},
  author={Caron, Mathilde and Touvron, Hugo and Misra, Ishan and J{\'e}gou, Herv{\'e} and Mairal, Julien and Bojanowski, Piotr and Joulin, Armand},
  booktitle={Proceedings of the IEEE/CVF international conference on computer vision},
  pages={9650--9660},
  year={2021}
}

@inproceedings{pal2024gamma,
  title={GAMMA-FACE: GAussian Mixture Models Amend Diffusion Models for Bias Mitigation in Face Images},
  author={Pal, Basudha and Kannan, Arunkumar and Kathirvel, Ram Prabhakar and O’Toole, Alice J and Chellappa, Rama},
  booktitle={European Conference on Computer Vision},
  pages={471--488},
  year={2024},
  organization={Springer}
}

@inproceedings{pal2024diversinet,
  title={DiversiNet: Mitigating Bias in Deep Classification Networks across Sensitive Attributes through Diffusion-Generated Data},
  author={Pal, Basudha and Roy, Aniket and Kathirvel, Ram Prabhakar and O’Toole, Alice J and Chellappa, Rama},
  booktitle={2024 IEEE International Joint Conference on Biometrics (IJCB)},
  pages={1--10},
  year={2024},
  organization={IEEE}
}

@inproceedings{dhar2021pass,
  title={Pass: protected attribute suppression system for mitigating bias in face recognition},
  author={Dhar, Prithviraj and Gleason, Joshua and Roy, Aniket and Castillo, Carlos D and Chellappa, Rama},
  booktitle={Proceedings of the IEEE/CVF International Conference on Computer Vision},
  pages={15087--15096},
  year={2021}
}

@article{schwemmer2020diagnosing,
  title={Diagnosing gender bias in image recognition systems},
  author={Schwemmer, Carsten and Knight, Carly and Bello-Pardo, Emily D and Oklobdzija, Stan and Schoonvelde, Martijn and Lockhart, Jeffrey W},
  journal={Socius},
  volume={6},
  pages={2378023120967171},
  year={2020},
  publisher={SAGE Publications Sage CA: Los Angeles, CA}
}

@inproceedings{siddiqui2022examination,
  title={An examination of bias of facial analysis based bmi prediction models},
  author={Siddiqui, Hera and Rattani, Ajita and Ricanek, Karl and Hill, Twyla},
  booktitle={Proceedings of the IEEE/CVF Conference on Computer Vision and Pattern Recognition},
  pages={2926--2935},
  year={2022}
}

@article{metz2025dissecting,
  title={Dissecting Human Body Representations in Deep Networks Trained for Person Identification},
  author={Metz, Thomas M and Hill, Matthew Q and Myers, Blake and Gandi, Veda Nandan and Chilakapati, Rahul and O'Toole, Alice J},
  journal={arXiv preprint arXiv:2502.15934},
  year={2025}
}

@book{cover1999elements,
  title={Elements of information theory},
  author={Cover, Thomas M},
  year={1999},
  publisher={John Wiley \& Sons}
}

@inproceedings{wang2022pose,
  title={Pose-guided feature disentangling for occluded person re-identification based on transformer},
  author={Wang, Tao and Liu, Hong and Song, Pinhao and Guo, Tianyu and Shi, Wei},
  booktitle={Proceedings of the AAAI conference on artificial intelligence},
  volume={36},
  number={3},
  pages={2540--2549},
  year={2022}
}

@inproceedings{li2023dc,
  title={Dc-former: Diverse and compact transformer for person re-identification},
  author={Li, Wen and Zou, Cheng and Wang, Meng and Xu, Furong and Zhao, Jianan and Zheng, Ruobing and Cheng, Yuan and Chu, Wei},
  booktitle={Proceedings of the AAAI Conference on Artificial Intelligence},
  volume={37},
  number={2},
  pages={1415--1423},
  year={2023}
}

@INPROCEEDINGS{pal2025quantitative,
  author={Pal, Basudha and Huang, Siyuan and Chellappa, Rama},
  booktitle={2025 IEEE International Joint Conference on Biometrics (IJCB)}, 
  title={A Quantitative Evaluation of the Expressivity of BMI, Pose and Gender in Body Embeddings for Recognition and Identification}, 
  year={2025},
  volume={},
  number={},
  pages={1-10},
  doi={10.1109/IJCB65343.2025.11411132}}

@inproceedings{nanduri2024template,
  title={Template-based multi-domain face recognition},
  author={Nanduri, Anirudh and Chellappa, Rama},
  booktitle={2024 IEEE International Joint Conference on Biometrics (IJCB)},
  pages={1--10},
  year={2024},
  organization={IEEE}
}

@inproceedings{kalka2019iarpa,
  title={Iarpa janus benchmark multi-domain face},
  author={Kalka, Nathan D and Duncan, James A and Dawson, Jeremy and Otto, Charles},
  booktitle={2019 IEEE 10th International Conference on Biometrics Theory, Applications and Systems (BTAS)},
  pages={1--9},
  year={2019},
  organization={IEEE}
}

@inproceedings{nanduri2025multi,
  title={Multi-Domain Biometric Recognition using Body Embeddings},
  author={Nanduri, Anirudh and Huang, Siyuan and Chellappa, Rama},
  booktitle={2025 IEEE 19th International Conference on Automatic Face and Gesture Recognition (FG)},
  pages={1--10},
  year={2025},
  organization={IEEE}
}

@article{wang2024yolov10,
  title={Yolov10: Real-time end-to-end object detection},
  author={Wang, Ao and Chen, Hui and Liu, Lihao and Chen, Kai and Lin, Zijia and Han, Jungong and Ding, Guiguang},
  journal={Advances in neural information processing systems},
  volume={37},
  pages={107984--108011},
  year={2024}
}

@inproceedings{fu2021unsupervised,
  title={Unsupervised pre-training for person re-identification},
  author={Fu, Dengpan and Chen, Dongdong and Bao, Jianmin and Yang, Hao and Yuan, Lu and Zhang, Lei and Li, Houqiang and Chen, Dong},
  booktitle={Proceedings of the IEEE/CVF conference on computer vision and pattern recognition},
  pages={14750--14759},
  year={2021}
}

@inproceedings{rajasegaran2022tracking,
  title={Tracking People by Predicting 3{D} Appearance, Location \& Pose},
  author={Rajasegaran, Jathushan and Pavlakos, Georgios and Kanazawa, Angjoo and Malik, Jitendra},
  booktitle={CVPR},
  year={2022}
}

@inproceedings{goel2023humans,
    title={Humans in 4{D}: Reconstructing and Tracking Humans with Transformers},
    author={Goel, Shubham and Pavlakos, Georgios and Rajasegaran, Jathushan and Kanazawa, Angjoo and Malik, Jitendra},
    booktitle={ICCV},
    year={2023}
}

@inproceedings{zhang2023diverse,
	title={Diverse embedding expansion network and low-light cross-modality benchmark for visible-infrared person re-identification},
	author={Zhang, Yukang and Wang, Hanzi},
	booktitle={Proceedings of the IEEE/CVF conference on computer vision and pattern recognition},
	pages={2153--2162},
	year={2023}
}
}

%

\begin{IEEEbiography} [{\includegraphics[width=1in,height=1.25in,clip,keepaspectratio]{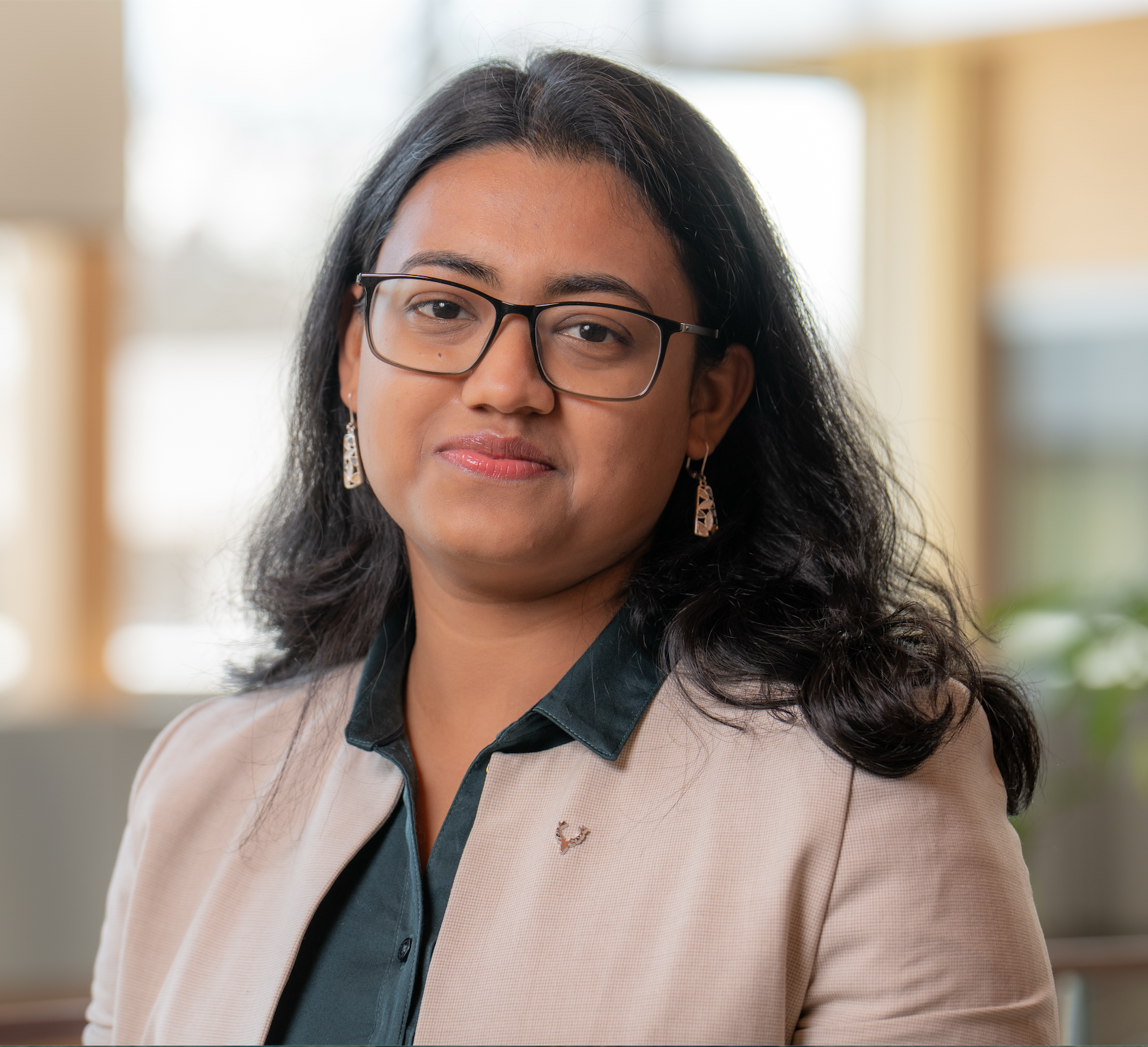}}]{Basudha Pal}  is a recent Ph.D. graduate in Electrical and Computer Engineering from Johns Hopkins University, where she was advised by Prof. Rama Chellappa. Her research focuses on trustworthy representation learning in computer vision, with applications in biometrics and medical imaging. She develops methods to quantify and refine correlations in deep learning models using mutual information estimation, data attribution, and generative modeling. During her doctoral research, she contributed to projects funded by the Intelligence Advanced Research Projects Activity (IARPA) and the National Institutes of Health (NIH), and worked on translational medical imaging systems at Johnson \& Johnson Innovative Medicine. Her work has been presented at leading venues including ECCV, MICCAI, IJCB, ISBI and RSNA. She previously received her B.Tech. degree in Electronics and Communication Engineering from Manipal Academy of Higher Education, India and was selected as an MITACS Globalink Research Fellow. 
\end{IEEEbiography}

\begin{IEEEbiography}[{\includegraphics[width=1in,height=1.25in,clip,keepaspectratio]{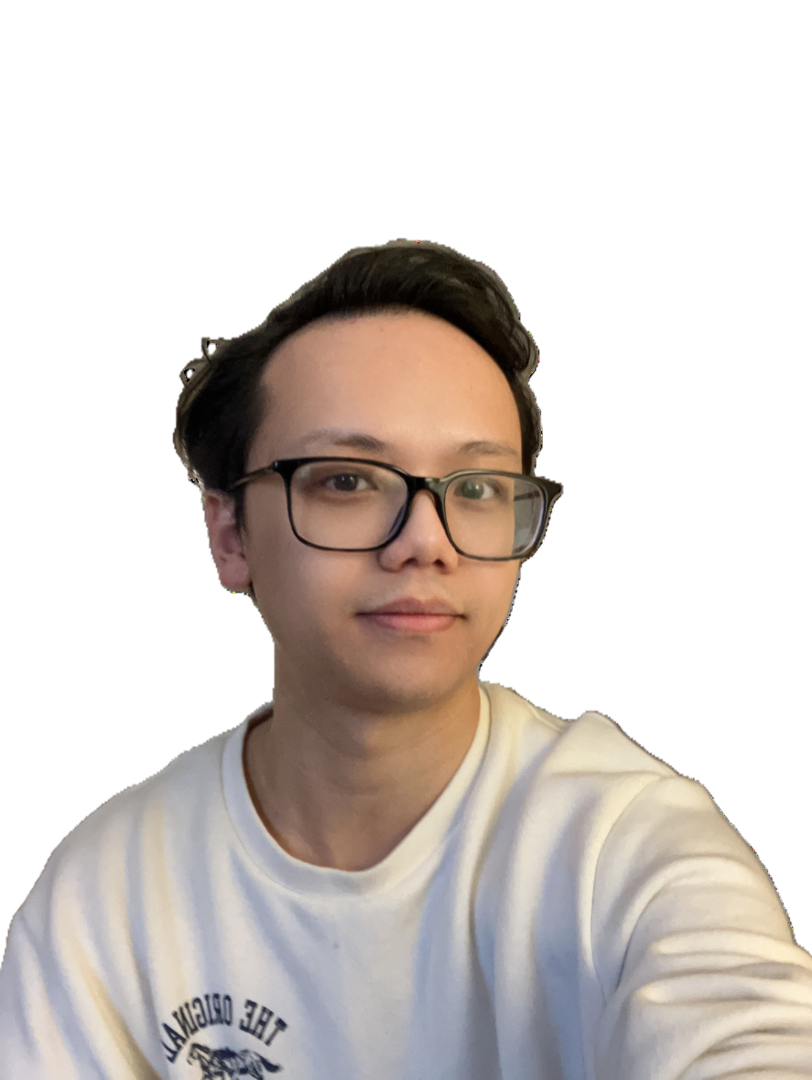}}]{Siyuan Huang} is currently a Ph.D. candidate in Computer Engineering at Johns Hopkins University, under the supervision of Prof. Rama Chellappa. He received the M.S.E. degree in Computer Engineering from Johns Hopkins University in 2024, and the M.S. degree in Computer Science from The George Washington University in 2020. His research interests include biometrics, person re-identification, gait recognition, multi-modal learning, computer vision, large language models, and agentic AI. His recent work focuses on unconstrained biometric recognition in challenging real-world environments, including video-image representation learning for cross-modal person re-identification, distillation-guided gait recognition for long-range human authentication, and body embedding evaluation for whole-body biometrics. His research has been deployed in IARPA BRIAR evaluations. He received the IJCB 2024 Best Student Paper Award, two IJCB 2025 Oral Presentations, two CVPR 2022 Oral Presentations, the AI2AI JHU + Amazon Initiative for Interactive AI Ph.D. Fellowship in 2026, and Adobe Research Funding in 2026. He has served as a reviewer for CVPR, NeurIPS, ECCV, ICCV, ICML, and IEEE TPAMI.
\end{IEEEbiography}

\begin{IEEEbiography}[{\includegraphics[width=1in,height=1.25in,clip,keepaspectratio]{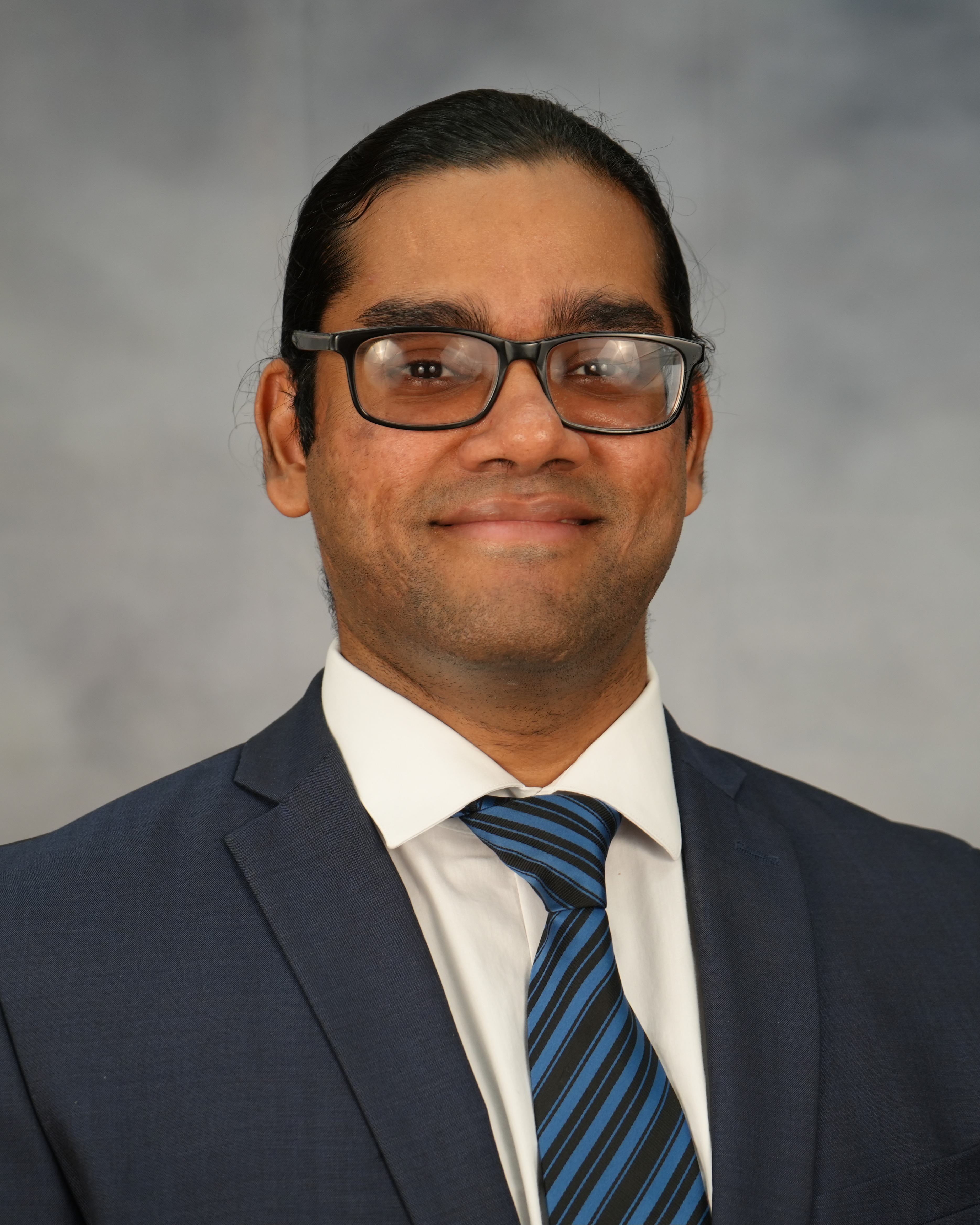}}]{Anirudh Nanduri}
received the B.Tech. degree from the Indian Institute of Technology Bombay, Mumbai, India, in 2015, and the Ph.D. degree in Electrical and Computer Engineering from the University of Maryland, College Park, MD, USA, in 2025. He was most recently a Postdoctoral Researcher with the University of Maryland. His research interests include cross-spectral biometric recognition, domain adaptation, gait analysis and 3D computer vision.
\end{IEEEbiography}

\begin{IEEEbiography}[{\includegraphics[width=1in,height=1.25in,clip,keepaspectratio]{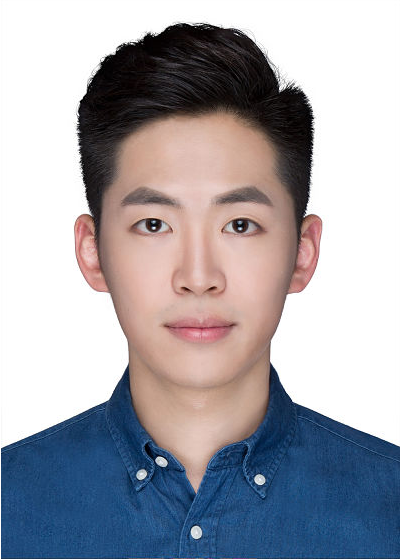}}]{Zhaoyang Wang}
is a Ph.D. candidate in Computer Science at Johns Hopkins University, advised by Prof. Rama Chellappa. As a member of the Artificial Intelligence for Engineering and Medicine (AIEM) Lab, his research focuses on computer vision and machine learning for human behavior understanding and medical applications. During his Ph.D., he has been involved in multiple projects funded by the Intelligence Advanced Research Projects Activity (IARPA), the Department of Defense (DoD), the National Institutes of Health (NIH), and the National Institute on Aging (NIA). His research has been published in top conferences and journals, including NeurIPS, ICCV, ICLR, MICCAI, WACV, IJCB, FG, IROS, IEEE Robotics and Automation Letters (RA-L), and Pattern Recognition (PR). He received the Hamlyn Prize from Imperial College London, where he earned his master’s degree in Medical Robotics and Image-Guided Intervention under the supervision of Prof. Guang-Zhong Yang.
\end{IEEEbiography}

\begin{IEEEbiography}[{\includegraphics[width=1in,height=1.25in,clip,keepaspectratio]{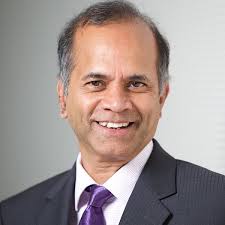}}]{Rama Chellappa}
Rama Chellappa is a Bloomberg Distinguished Professor in the Departments of Electrical and Computer Engineering and Biomedical Engineering at Johns Hopkins University (JHU). His research interests span artificial intelligence, computer vision, machine learning, signal, image, and video processing. Professor Chellappa’s scholarship has been recognized with numerous prestigious honors. Recent awards include the 2025 Azriel Rosenfeld Lifetime Achievement Award and the 2023 Distinguished Researcher Award from the IEEE Computer Society’s PAMI Technical Committee, as well as the 2024 Edwin H. Land Medal from Optica and the 2020 IEEE Jack S. Kilby Medal for Signal Processing. He is also the recipient of the 2012 K. S. Fu Prize from the International Association of Pattern Recognition; the Society, Technical Achievement, and Meritorious Service Awards from the IEEE Signal Processing Society; the Technical Achievement and Meritorious Service Awards from the IEEE Computer Society; and the Leadership Award from the IEEE Biometrics Council. He is an elected member of the U.S. National Academy of Engineering and a Foreign Fellow of the Indian National Academy of Engineering. He is also a Fellow of AAAI, AAAS, ACM, AIMBE, IAPR, IEEE, NAI, Optica, and the Washington Academy of Sciences. Professor Chellappa holds nine U.S. patents.
\end{IEEEbiography}





\end{document}